\pdfoutput=1
\documentclass{article}

\usepackage{microtype}
\usepackage{graphicx}
\usepackage{booktabs}
\usepackage{hyperref}

\usepackage[preprint]{icml2027}

\usepackage[utf8]{inputenc}
\usepackage[T1]{fontenc}
\usepackage{url}
\usepackage{xurl}  %
\usepackage{amsmath}
\usepackage{amsfonts}
\usepackage{amssymb}
\usepackage{nicefrac}
\usepackage{xcolor}
\usepackage{longtable}
\usepackage{enumitem}  %

\newcommand{\pizero}{$\pi_0$}
\newcommand{\pihalf}{$\pi_{0.5}$}

\newcommand{\nPolicies}{44}
\newcommand{\nModels}{41}           %
\newcommand{\nAgents}{three}        %
\newcommand{\nTests}{212}
\newcommand{\nTrials}{395{,}092}
\newcommand{\nLinked}{25}           %
\newcommand{\nBothGroups}{13}       %
\newcommand{\firstRelease}{December 2022}
\newcommand{\lastRelease}{October 2026}

\newcommand{\shrinkShare}{82\%}     %
\newcommand{\cvLossShared}{0.3585}
\newcommand{\linkLossShared}{0.3934}
\newcommand{\linkLossPerBench}{0.3942}
\newcommand{\linkLossPerSetting}{0.3932}

\newcommand{\stableShare}{1\%}      %
\newcommand{\stableTrials}{3{,}959}

\newcommand{\eloPerYear}{171}       %
\newcommand{\eloPerYearRound}{170}
\newcommand{\yearScore}{73\%}       %
\newcommand{\paceBefore}{70}
\newcommand{\paceSince}{730}

\newcommand{\avgPaceBefore}{110}
\newcommand{\avgPaceSince}{610}

\newcommand{\sotaOdds}{18}
\newcommand{\oddsDoubling}{8.5}
\newcommand{\recentRatio}{five}
\newcommand{\benchEloPerYear}{372}
\newcommand{\benchEloPerYearRound}{370}   %
\newcommand{\benchOdds}{8.5}
\newcommand{\sotaBacktest}{116}
\newcommand{\sotaAboveBand}{five}
\newcommand{\dojoMedian}{2250}      %
\newcommand{\dojoNinety}{2758}      %
\newcommand{\recentHalfYear}{280}   %
\newcommand{\dojoRecordRelease}{1931}     %
\newcommand{\dojoRecordNow}{2195}         %
\newcommand{\dojoPace}{880}               %
\newcommand{\dojoNinetyDate}{June 2027}
\newcommand{\dojoNinetyLo}{April 2027}
\newcommand{\dojoNinetyHi}{September 2027}
\newcommand{\usedUpLo}{7}                 %
\newcommand{\usedUpHi}{21}                %
\newcommand{\forecastMedian}{January 2027}
\newcommand{\forecastNinety}{September 2027}   %
\newcommand{\forecastNinetyDate}{September 2027}
\newcommand{\forecastNinetyLo}{April 2027}
\newcommand{\forecastNinetyHi}{January 2028}
\newcommand{\needHi}{3480}
\newcommand{\nextTrendDate}{July 2027}
\newcommand{\nextTrendLo}{2240}
\newcommand{\nextTrendHi}{3070}
\newcommand{\nextTrendScore}{56\%}
\newcommand{\nextTrendMonths}{eight}
\newcommand{\forecastHistLo}{January 2027}      %
\newcommand{\forecastHistHi}{April 2028}        %

\icmltitlerunning{How Much Has Embodied AI Progressed?}

\begin{document}

\twocolumn[
  \icmltitle{How Much Has Embodied AI Progressed? A Meta-Analysis on\\Robot Manipulation Benchmarks vs. VLA, WAM, and LLM Agents}

  \icmlsetsymbol{equal}{*}
  \icmlsetsymbol{corresponding}{$\dagger$}

  \begin{icmlauthorlist}
    \icmlauthor{Gang Yang}{equal,aff1}
    \icmlauthor{Nigel Jian Wei Goh}{equal,aff1}
    \icmlauthor{Shuo Sun}{corresponding,aff2,aff3}
    \icmlauthor{Christina Dao Wen Lee}{aff1}
    \icmlauthor{Xinru Wang}{aff2,aff3}
    \\
    \icmlauthor{Alok Prakash}{aff2,aff3}
    \icmlauthor{Chee Leong Teo}{aff1}
    \icmlauthor{Marcelo H. Ang Jr.}{aff1}
    \icmlauthor{Thomas W. Malone}{aff3,aff4}
  \end{icmlauthorlist}

  \icmlaffiliation{aff1}{
    Advanced Robotics Centre,
    National University of Singapore,
    Singapore
  }

  \icmlaffiliation{aff2}{
    Mens, Manus and Machina (M3S),
    Singapore-MIT Alliance for Research and Technology (SMART),
    Singapore
  }

  \icmlaffiliation{aff3}{
    MIT Center for Collective Intelligence,
    Massachusetts Institute of Technology,
    Cambridge, MA, USA
  }

  \icmlaffiliation{aff4}{
    Sloan School of Management,
    Massachusetts Institute of Technology,
    Cambridge, MA, USA
  }

  \icmlcorrespondingauthor{Shuo Sun}{shuosun@mit.edu}

  \icmlkeywords{Robot Manipulation, Evaluation, Benchmarks, VLA, WAM, LLM Agent}

  {\centering
    \vskip 0.1in
    \href{https://lalayang123456-ctrl.github.io/roboelo-web/}{\textbf{Project Website}}\par
  } %

  \vskip 0.1in
  \begin{center}
  \includegraphics[width=\textwidth]{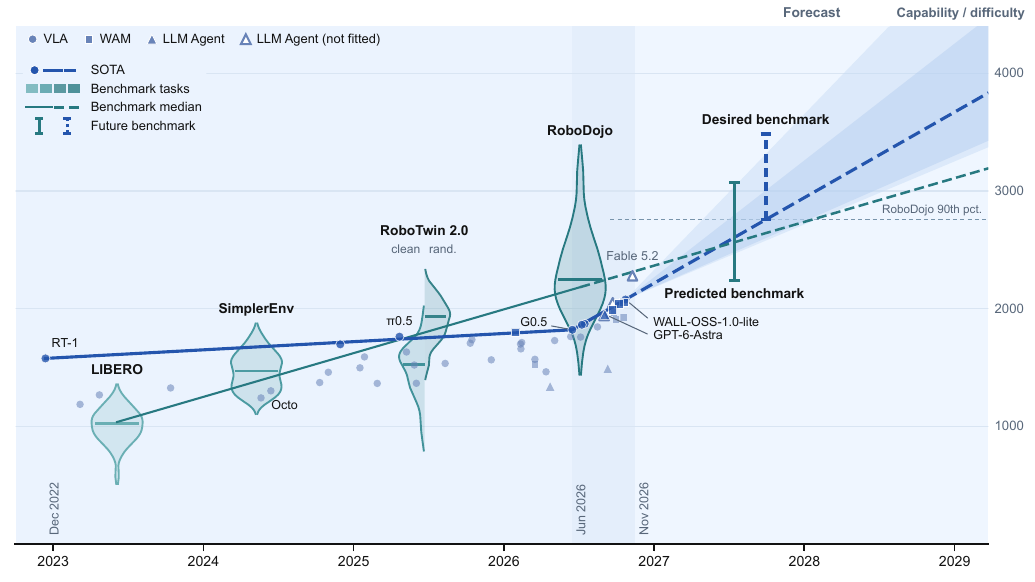}
  \captionof{figure}{\textbf{Policy capability and benchmark difficulty on a common Elo scale, 2022--2026.} Robot policies have improved about ten times faster since June 2026, and benchmarks saturate within months.
  Policies (markers: general capability $\mu$ at release; dark blue: SOTA) and benchmark tasks (green violins, at release) share one unified scale, in which a policy expects a 50\% success rate on a task of equal Elo rating (Table~\ref{tab:selected}).
  The SOTA trend line bends at G0.5~\citep{galaxea2026g05} (June 2026); blue dashed line: forecast at the pace since then (bands: 95\% CIs).
  Future benchmarks: the predicted benchmark based on the observed trend (solid green, July 2027, 10th--90th percentile) and the desired benchmark, the difficulty range needed to last a year once RoboDojo~\citep{chen2026robodojo} is saturated (the best policy expects $\geq$\,50\% on 90\% of its tasks; dashed, September 2027).
  Hollow triangles: LLM Agents, placed from their mean scores listed on the RoboDojo leaderboard (contributed by Anthropic; \citep{anthropic2026fable52}); not fitted and not used in the trend lines as no per-task results are published.}
  \label{fig:teaser}
\end{center}

  \vskip 0.3in
]

\printAffiliationsAndNotice{
  \textsuperscript{*}Equal contribution.
  \textsuperscript{$\dagger$}Corresponding author.
}

\begin{abstract}
Robot manipulation policies, from vision-language-action models (VLAs) and world-action models (WAMs) to, most recently, large language model (LLM) agents, now improve within months.
However, each benchmark separates them only over a narrow range: older benchmarks saturate, while on newer ones most policies score in the single digits, so policies evaluated on different benchmarks cannot be compared and the pace of progress cannot be measured.
We place policy capabilities and task difficulties from different benchmarks on one Elo scale, linking the benchmarks through the policies they share, as test linking does for exams in educational measurement; on this scale, a policy's odds of success on a task are the ratio of its capability to the task's difficulty.
Applied to \nPolicies{} policies released between \firstRelease{} and \lastRelease{} on LIBERO, SimplerEnv, RoboTwin~2.0 and RoboDojo (\nTests{} test cases, \nTrials{} trials), the scale predicts a policy's results on a benchmark hidden from the fit with an error of 6.8 points in mean score, against 11.9 for the per-task average.
On this unified scale, today's best policy is about \sotaOdds{} times as capable as RT-1.
On average, policy capability doubled every \oddsDoubling{} months, and since June 2026 the best policy has improved about ten times as fast as before.
Meanwhile, the difficulty of new benchmarks grew about \benchOdds{}-fold a year (95\% CI: 2--36): the median task of today's hardest benchmark, RoboDojo, is about 1{,}200 times as difficult as LIBERO's.
Benchmarks that were hard at release saturated \usedUpLo{} to \usedUpHi{} months later, and every benchmark except RoboDojo is already saturated.
If the pace since June 2026 holds, the best policy will saturate RoboDojo around \forecastNinety{}, and the tasks that remain hard, open-ended tasks in randomized scenes, are where the next benchmark should build.
\end{abstract}

\section{Introduction}
\label{sec:intro}

In robotic manipulation tasks, embodied AI policies are improving quickly.
Since the end of 2022, new vision-language-action models (VLAs) \citep{kim2024openvla,black2024pi0}, which turn camera images and a language instruction into robot actions with a pretrained vision-language model, world-action models (WAMs) \citep{ye2026dreamzero}, which also predict how the scene will evolve, and LLM Agents \citep{zhang2026unexpected}, which drive the robot through a fixed harness, appear almost every month.
In the meantime, new and more challenging benchmarks for evaluating these policies are developing at a yearly pace.
Yet, a benchmark can tell policies apart only within a limited range of capability, much as a ruler measures only up to its own length.
Once the best policies achieve nearly 100\% success rate, as observed on the LIBERO benchmark~\citep{liu2023libero}, the benchmark can no longer separate them.
Newer benchmarks are harder and are able to separate the most capable policies, but in the initial stages, most policies often only achieve single-digit success rates on them, providing little separation either.
The faster the field advances, the sooner each benchmark saturates and is superseded by the next one.
This leaves two basic questions unanswered: how to compare policies that share no benchmark, and how to measure how fast the field is advancing.
The first arises because policies end up evaluated on different benchmarks, so two policies often have none in common. The second arises because no single benchmark spans the whole course of progress.

Both of these questions can be solved by putting all the benchmarks on a single scale. This adopts a similar idea used in educational measurement where tests of different difficulties are put together on one scale.
Test linking places scores from different test forms on a common scale, using the examinees or items the forms share. Vertical scaling extends this to tests written for different school grades, producing a single scale on which student growth can be tracked \citep{kolen2014test}.
AI evaluation has borrowed these ideas to rate models together with the test items they face \citep{rodriguez2021evaluation,sun2025agielo} and to stitch language-model benchmarks into a single scale for tracking long-run trends \citep{ho2025rosetta}.
Robot policy evaluation, by contrast, has focused on new platforms that run new trials \citep{li2024simpler,atreya2025roboarena,duan2026robocolosseum}.
No work has yet linked the existing robot manipulation benchmarks, and the pace of progress in robot manipulation has not been measured.

\textbf{One Elo scale across embodied AI benchmarks.}
We develop a simple statistical framework that stitches existing per-task results onto one scale (Figure~\ref{fig:teaser}).
Each policy has a capability and each task a difficulty, and the odds that the policy succeeds on the task are the ratio of the two.
We express both on the Elo scale familiar from chess~\citep{elo1978rating}, on which a 400-point lead means an expected score of 91\% on any task.
All per-task results from all benchmarks are fitted together, and a policy evaluated on two benchmarks links them.
The result is a unified scale that holds for every policy and every task: two policies can be compared even if they never ran the same benchmark, and the difficulty of each benchmark can be read off the same axis (Section~\ref{sec:method}).

We apply the framework to \nPolicies{} policies released between \firstRelease{} and \lastRelease{}: \nModels{} influential vision-language-action and world-action models, and \nAgents{} LLM Agents that drive the robot through a fixed harness. They are evaluated on four benchmarks: LIBERO, SimplerEnv \citep{li2024simpler}, RoboTwin~2.0 \citep{chen2025robotwin2} and RoboDojo \citep{chen2026robodojo}.
The policies start with RT-1 \citep{brohan2022rt1}, the precursor of VLA models \citep{zitkovich2023rt2}.
The data combine public leaderboards with our own evaluations and contain \nTrials{} trials on \nTests{} test cases.
On benchmarks hidden from the fit, the scale predicts a policy's mean score with an error of 6.8 points, against 11.9 for the per-task average (Section~\ref{sec:data}).

We use the scale to answer four questions about progress (Section~\ref{sec:results}):
\begin{itemize}[leftmargin=*, topsep=2pt, itemsep=2pt, parsep=0pt]
  \item \textbf{Where do policies stand?} Today's best policy, WALL-OSS-1.0-lite, is about \sotaOdds{} times as capable as RT-1: on a task where RT-1 succeeds about half the time, it would score 95\%. Six of the eight best policies are WAMs released in September and October 2026, and an LLM Agent with no robot training ranks sixth (Section~\ref{sec:stand}).
  \item \textbf{How fast do policies improve?} From December 2022 to October 2026, policy capabilities have risen by about \eloPerYearRound{}~Elo points per year, multiplying capability by about $2.7\times$ a year: a policy released \oddsDoubling{} months later is, on average, twice as capable. Progress is speeding up: since G0.5 in June 2026, the best policy has risen by about \paceSince{}~Elo points per year, about ten times its earlier pace, and the speed-up also shows in raw scores, without the scale (Section~\ref{sec:progress}).
  \item \textbf{Where do benchmarks stand?} Each new benchmark is harder than the last: benchmark difficulty rises by about \benchEloPerYearRound{}~Elo points per year, about twice as fast as the average policy improves, and RoboDojo's median task is about 1{,}200 times as difficult as LIBERO's. Every benchmark except RoboDojo is already saturated (Section~\ref{sec:benchstand}).
  \item \textbf{How fast do benchmarks saturate?} Benchmarks that were hard at release saturated within \usedUpLo{} to \usedUpHi{} months. Since G0.5 the best policy has improved faster than benchmarks have hardened; if this pace holds, RoboDojo, the hardest benchmark, will saturate around \forecastNinety{}. The tasks that remain hard, open-ended tasks in randomized scenes, are where a next benchmark should build, and it would be needed before then to keep the field's momentum (Section~\ref{sec:left}).
\end{itemize}

\section{Findings}
\label{sec:results}

This section reports what one Elo scale over four benchmarks shows: where policies stand today (Section~\ref{sec:stand}), how fast they improve (Section~\ref{sec:progress}), where benchmarks stand (Section~\ref{sec:benchstand}), and how fast they saturate (Section~\ref{sec:left}).
    Figure~\ref{fig:teaser} gives the overview, and Section~\ref{sec:method} describes how the scale is built.

\subsection{Data and Validation}
\label{sec:data}

\paragraph{Benchmarks and policies.}
We combine four widely used simulation benchmarks for manipulation policies, released from 2023 to 2026 (Table~\ref{tab:data}): LIBERO and SimplerEnv use one robot arm, and RoboTwin~2.0 and RoboDojo use two.
We rate \nPolicies{} policies spanning across VLAs, WAMs and LLM Agents, released between \firstRelease{} and \lastRelease{}, as summarized in Table~\ref{tab:policies}.
We chose them to represent the field, not to include every policy: the top policies on each benchmark, influential policies that others build on, successive versions of the same family, and policies tested on several benchmarks, which link the benchmarks together.
Among the 395,092 trials we collected, 74\% come from public leaderboards, 3\% from one published paper and 23\% were reproduced runs for this study under each benchmark's official protocol. For each policy's trials on a benchmark task, a full success scores 1, and a failure earns partial credit for its progress when the benchmark records it (Appendix~\ref{app:data}).
Of the \nPolicies{} policies, \nLinked{} ran at least two benchmarks and so link them, and \nBothGroups{} ran both a one-arm and a two-arm benchmark.

\begin{table}[ht]
  \centering
  \setcounter{table}{0}%
  \caption{\textbf{Benchmark statistics.} Test cases are tasks under given conditions. RoboDojo's policies include the \nAgents{} LLM Agents.}
  \label{tab:data}
  \small
  \setlength{\tabcolsep}{3.5pt}
  \begin{tabular}{@{}llcrrr@{}}
    \toprule
    Benchmark & Released & Arms & Test cases & Policies & Trials \\
    \midrule
    LIBERO       & 2023-06 & 1 & 40  & 21 & 40{,}500  \\
    SimplerEnv   & 2024-05 & 1 & 18  & 15 & 41{,}232  \\
    RoboTwin 2.0 & 2025-06 & 2 & 100 & 17 & 160{,}000 \\
    RoboDojo     & 2026-07 & 2 & 54  & 27 & 153{,}360 \\
    \midrule
    Total        &         &   & 212 & 44 & 395{,}092 \\
    \bottomrule
  \end{tabular}
\end{table}
\setcounter{table}{2}%

\begin{figure}[ht]
  \centering
  \includegraphics[width=\linewidth]{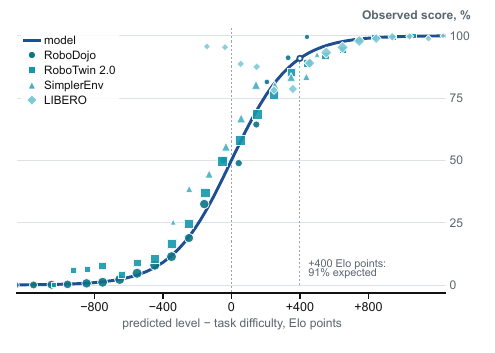}
  \caption{\textbf{Predicted against observed scores on benchmarks hidden from the fit.} Each of \nLinked{} policies has all its results on one benchmark hidden and predicted from its other benchmarks (61 hidden policy–benchmark pairs, 3,214 policy–task cells). Curve: the model's expected score; dots: trial-weighted means in bins of 100~Elo points (size: trials).}
  \label{fig:hidden}
\end{figure}

\begin{table*}[t]
  \centering
  \setcounter{table}{1}%
  \caption{\textbf{Representative policies vs. benchmark tasks of matching difficulty.} The top eight policies and eight more spread over the ranking, by rank among all \nPolicies{} in general capability $\mu$ (95\% CIs of $\mu$ in Table~\ref{tab:policies}). Task at this level: of the test cases the policy ran within 40~Elo points of $\mu$ and scored 35--65\% on, the one whose difficulty $\beta$ is closest to $\mu$; score: the policy's observed mean score on it. Ran: L=LIBERO, S=SimplerEnv, T=RoboTwin~2.0, D=RoboDojo.}
  \label{tab:selected}
  \small
  \setlength{\tabcolsep}{3pt}%
  \begin{tabular}{@{}rlllllll@{\hspace{3pt}}c@{}}
    \toprule
    \# & Policy & Type & Released & Ran & $\mu$ & Task at this level & $\beta$ & Score \\
    \midrule
    1  & WALL-OSS-1.0-lite~\citep{xsquare2026wallosslite} & VLA       & 2026-10 & D   & 2078 & pour balls into vase (D)                & 2078 & 52\% \\
    2  & PrismWAM$^{*}$~\citep{xin2026prismwam} & WAM       & 2026-10 & TD  & 2056 & store laptop and headphones (D)         & 2069 & 58\% \\
    3  & VPP2~\citep{guo2026vpp2} & WAM       & 2026-10 & D   & 2042 & fill pen holder (D)                     & 2040 & 50\% \\
    4  & PatchWAM~\citep{wang2026patchwam} & WAM       & 2026-09 & T   & 1990 & turn switch (T, rand.)             & 1988 & 36\% \\
    5  & InternW0-$\Delta$~\citep{miao2026internw0} & WAM       & 2026-09 & D   & 1970 & organize table (D)                      & 1952 & 60\% \\
    6  & GPT-6-Astra~\citep{openai2026gpt6astra} & LLM Agent & 2026-09 & D   & 1953 & organize table (D)                      & 1952 & 40\% \\
    7  & PrismWAM~\citep{xin2026prismwam} & WAM       & 2026-10 & TD  & 1928 & beat block hammer (T, rand.)       & 1927 & 48\% \\
    8  & EWAM~\citep{wang2026ewam} & WAM       & 2026-09 & TD  & 1909 & lift pot (T, rand.)                & 1906 & 48\% \\
    15 & $\pi_{0.5}$~\citep{pi2025pi05} & VLA       & 2025-04 & LTD & 1762 & press stapler (T, rand.)           & 1760 & 63\% \\
    19 & Xiaomi-R0~\citep{cai2026xiaomir0} & VLA       & 2026-02 & LSTD & 1713 & stack two bowls (T, rand.)        & 1717 & 53\% \\
    22 & CogACT~\citep{li2024cogact} & VLA       & 2024-11 & S   & 1698 & put carrot on plate (S)                 & 1675 & 54\% \\
    26 & RT-1~\citep{brohan2022rt1} & VLA       & 2022-12 & S   & 1578 & put apple in drawer (S)                 & 1581 & 53\% \\
    31 & GR00T N1.5~\citep{nvidia2025gr00tn1} & VLA       & 2025-05 & LS  & 1522 & put eggplant in basket (S)              & 1507 & 58\% \\
    35 & $\pi_0$~\citep{black2024pi0} & VLA       & 2024-10 & LTD & 1460 & stack two blocks (T, clean)             & 1480 & 42\% \\
    41 & OpenVLA~\citep{kim2024openvla} & VLA       & 2024-06 & LS  & 1302 & put bowl in top drawer (L)              & 1276 & 50\% \\
    43 & Octo~\citep{octo2024octo} & VLA       & 2024-05 & S   & 1242 & close drawer (S)                        & 1251 & 52\% \\
    \bottomrule
  \end{tabular}
\end{table*}

\paragraph{Hidden-benchmark validation.}
Before using the scale, we test whether it predicts results it was not fitted on. For each of the \nLinked{} policies that ran at least two benchmarks, we hide all of its results on one benchmark, refit, and predict them from its other benchmarks.
Over the 61 hidden cases, the predicted mean score is off by 6.8 points, against 11.9 for using the task average (the other policies' mean on the same tasks); the scale predicts more accurately on all four benchmarks and orders two hidden policies correctly 71--85\% of the time (Appendix~\ref{sec:reliability}).
Beyond mean scores, the scale also predicts the score on each task: grouped by the predicted gap between a policy's level and a task's difficulty, the observed scores differ from the model's expected score by 2.9 points on average (Figure~\ref{fig:hidden}).

\subsection{Where Do Policies Stand?}
\label{sec:stand}

\paragraph{Interpreting policies' capability by benchmark tasks of the same difficulty level.}
Table~\ref{tab:selected} matches each policy to a task whose difficulty lies near its general capability ($\pm40$~Elo points), so a policy at this level has an expected score of around 50\% (observed median is 53\%) on it.
Today's best policy, WALL-OSS-1.0-lite~\citep{xsquare2026wallosslite}, as of October 2026, scores about half on RoboDojo tasks such as pouring balls into a vase (52\%) or playing tic-tac-toe (50\%).
The best policy of 2022, RT-1, is at the level of putting an apple into a drawer in SimplerEnv (53\%), a task on which today's best policy would score 95\%.
The two RoboDojo tasks are clearly harder for RT-1: they lie about 500~Elo points above it, and RT-1 would score only about 5\% on them.
The 500-Elo points gap between WALL-OSS-1.0-lite and RT-1 makes today's best policy about \sotaOdds{} times as capable (Equation~\ref{eq:odds}).

\paragraph{WAMs and an LLM Agent hold most of the top places.}
Six of the eight best policies are WAMs released between September and October 2026 (PrismWAM$^{*}$, \citealp{xin2026prismwam}; VPP2, \citealp{guo2026vpp2}; PatchWAM, \citealp{wang2026patchwam}; InternW0-$\Delta$, \citealp{miao2026internw0}; PrismWAM; and EWAM, \citealp{wang2026ewam}); first place goes to a VLA, WALL-OSS-1.0-lite, and sixth to an LLM Agent, GPT-6-Astra~\citep{openai2026gpt6astra}.
The LLM Agents get no robot training, yet they move fast: GPT-5.5 (April 2026; \citealp{openai2026gpt55}) ranks 39th, while GPT-6-Astra, five months later and with the same harness, ranks sixth.
Later still, Claude Fable 5.2 (November 2026) scores 65\% on RoboDojo as reported by Anthropic \citep{anthropic2026fable52}, above the 42\% of today's best policy, WALL-OSS-1.0-lite; it is shown in Figure~\ref{fig:teaser} but not fitted.
Their capabilities also have a different profile.
Compared with the scale's prediction, the agents score 12 points higher on RoboDojo's Open tasks and 4 points higher on its Memory tasks, but 7 points lower on Precision and 9 points lower on Long-Horizon tasks; for the other policies, these differences stay below one point.
LLM Agents thus bring capability on open-ended tasks but not yet the precise control of learned policies, although three agents on one benchmark are too few for a firm conclusion.

\subsection{How Fast Do Policies Improve?}
\label{sec:progress}

\begin{figure*}[t]
  \centering
  \includegraphics[width=\linewidth]{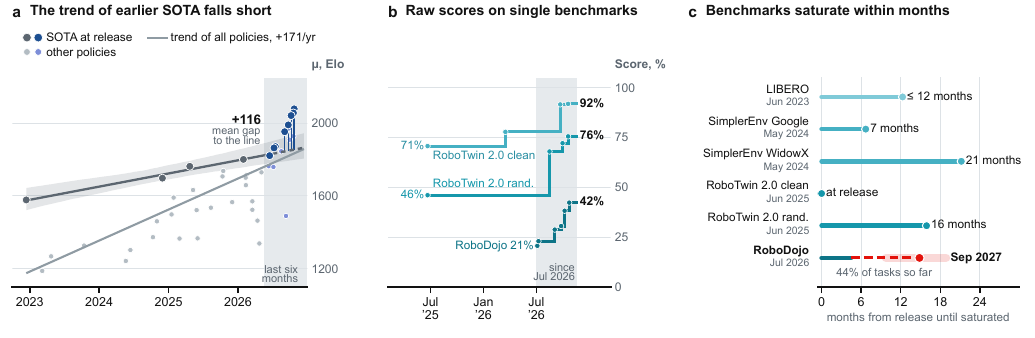}
  \caption{\textbf{Growth of the SOTA policies, and how fast benchmarks saturate.} (a)~A line through the four SOTA policies before the last six months (dark grey; band: 95\% CI, tasks resampled; $R^2 = 0.98$), and extrapolated (dashed): the eight later SOTA policies (dark blue) lie above it (vertical lines; $+\sotaBacktest{}$~Elo points on average), and \sotaAboveBand{} of them outside the 95\% CI; compared to the trend of all \nPolicies{} policies (light grey line; 95\% CI: 116--238; $R^2 = 0.55$). (b)~Best mean score so far on single benchmarks, among policies by release date; shaded: since July 2026. (c)~Months until the best policy scores at least 50\% on 90\% of a benchmark's tasks; red: RoboDojo's forecast at the pace since G0.5 (95\% CI).}
  \label{fig:progress}
\end{figure*}

\begin{figure*}[t]
  \centering
  \includegraphics[width=\linewidth]{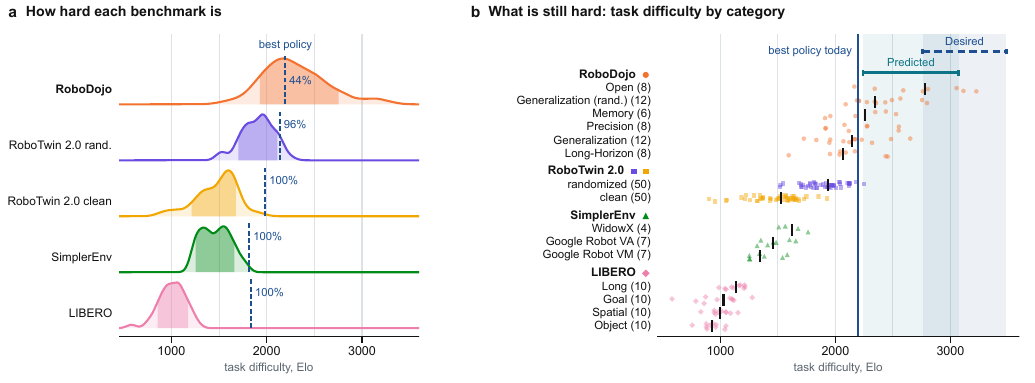}
  \caption{\textbf{Where benchmarks stand, and what remains hard.} (a)~Distribution of task difficulty on each benchmark (darker: the middle 80\% of tasks); dashed: the best policy's level on it today, with the share of tasks on which it expects to succeed at least half the time. (b)~Task difficulty by category; bars: medians; blue line: today's best level on RoboDojo; teal and dashed blue: the predicted and desired benchmarks of Figure~\ref{fig:teaser}.}
  \label{fig:usedup}
\end{figure*}

\paragraph{A policy released \oddsDoubling{} months later has, on average, twice the odds of success.}
A straight line through all \nPolicies{} policies (Figure~\ref{fig:progress}a) rises by \eloPerYear{}~Elo points per year (95\% CI: 116--238; $R^2 = 0.55$); on a typical task, this lifts an expected score of 50\% to about \yearScore{} within a year.
The growth holds on each benchmark on its own (115 to 278~Elo points per year) and under all 60 model settings of our robustness checks (Appendix~\ref{app:trend}).

\paragraph{Progress is speeding up.}
In the last six months the best policy gained about \recentHalfYear{}~Elo points, \recentRatio{} times its average six-month gain since 2022.
The growth rate accelerated after G0.5 (June 2026; Figure~\ref{fig:teaser}, Appendix~\ref{app:trend}): the SOTA rose by about \paceBefore{}~Elo points per year before it and \paceSince{} since, ten times as fast.
The field as a whole speeds up too: a line through all policies released before G0.5 rises by about \avgPaceBefore{}~Elo points per year, and through those released since, by about \avgPaceSince{}.
Two further checks point the same way.
First, a straight line through the SOTA policies released before May 2026 falls short of the eight SOTA of the last six months by \sotaBacktest{}~Elo points on average; seven of the eight lie above it, and \sotaAboveBand{} above its 95\% CI (Figure~\ref{fig:progress}a); lines fitted to all policies before earlier cut-offs fall short in the same way, and a curve with no kink also fits better than a line (Appendix~\ref{app:trend}).
Second, the speed-up shows in raw scores, without the scale: on RoboTwin~2.0 randomized, the best mean score stayed at 46\% for more than a year after the benchmark's release and then rose to 76\% between August and October 2026, and on RoboDojo it rose from 21\% to 42\% within four months of its release (Figure~\ref{fig:progress}b).

\subsection{Where Do Benchmarks Stand?}
\label{sec:benchstand}%

\paragraph{Each new benchmark gets harder and spreads wider.}
The median task Elo rating is 1024 on LIBERO, 1471 on SimplerEnv, 1524 on RoboTwin~2.0 clean, 1932 on RoboTwin~2.0 randomized and \dojoMedian{} on RoboDojo, in the order of their release (violins in Figure~\ref{fig:teaser}; Figure~\ref{fig:usedup}a).
Over three years, the median task rose by 1226~Elo points from LIBERO (June 2023) to RoboDojo (July 2026), making RoboDojo's median task about 1{,}200 times as difficult: a policy with an expected score of 50\% on LIBERO's median task would score below 0.1\% on RoboDojo's.
On average, benchmarks rise by about \benchEloPerYear{}~Elo points per year (green line in Figure~\ref{fig:teaser}), about twice the pace of the average policy (light gray line in Figure~\ref{fig:progress}a).
Newer benchmarks also spread their tasks wider: the middle 80\% of tasks (10th to 90th percentile of difficulty) spans about 320~Elo points on LIBERO, 410 on SimplerEnv, 470 on RoboTwin~2.0 clean, 420 on RoboTwin~2.0 randomized (720 for the two together) and 830 on RoboDojo (darker in Figure~\ref{fig:usedup}a).

\paragraph{All benchmarks except RoboDojo are already saturated.}
We call a benchmark \emph{saturated} when the policy with the highest level on it has an expected score of at least 50\% on 90\% of its tasks, since only 10\% of its tasks then lie above that level; it can no longer separate the best policies.
Figure~\ref{fig:usedup}a shows where the best policy on each benchmark stands.
On LIBERO, SimplerEnv and RoboTwin~2.0, the best policy already succeeds at least half the time on almost every task (96--100\%), making them saturated.
On RoboDojo, the best policy (WALL-OSS-1.0-lite) succeeds at least half the time on only 44\% of the tasks, leaving it not yet saturated.

\subsection{How Fast Do Benchmarks Saturate?}
\label{sec:left}

\paragraph{Benchmarks saturate within months.}
Benchmarks that were hard at their release saturated \usedUpLo{} to \usedUpHi{} months later: LIBERO lasted at most 12 months; SimplerEnv's Google Robot tasks lasted 7 months, and its WidowX tasks lasted 21 months; RoboTwin~2.0 clean was saturated at its release by $\pi_{0.5}$~\citep{pi2025pi05}, its randomized tasks lasted 16 months (Figure~\ref{fig:progress}c).
In the four months after RoboDojo's release, the share of its tasks on which the best policy expects to succeed at least half the time rose from 11\% (G0.5) to 44\% (WALL-OSS-1.0-lite).

\paragraph{RoboDojo is likely to saturate around \forecastNinety{}.}
Since G0.5 (June 2026), the best policy has risen at a pace of about \paceSince{}~Elo points per year, faster than the benchmarks have risen, so the gap that a new benchmark opens now closes faster than before (Figure~\ref{fig:teaser}).
At this pace, it reaches RoboDojo's median task difficulty in \forecastMedian{} and its 90th percentile, at which RoboDojo is saturated, in \forecastNinetyDate{} (95\% CI: \forecastNinetyLo{}--\forecastNinetyHi{}; Figure~\ref{fig:progress}c).
Two other estimates agree.
The best level on RoboDojo itself, $\mu_m + \gamma_{ms}$, which includes gains specific to RoboDojo, rose from \dojoRecordRelease{}~Elo points at RoboDojo's release in July 2026 to \dojoRecordNow{}~Elo points for WALL-OSS-1.0-lite, about \dojoPace{}~Elo points per year, which gives \dojoNinetyDate{}; the lifetimes of earlier benchmarks give \forecastHistLo{} to \forecastHistHi{}.
If benchmarks keep rising at their own pace, the next one would arrive around \nextTrendDate{}, one average release gap after RoboDojo, with tasks from about \nextTrendLo{} to \nextTrendHi{}~Elo points (the predicted benchmark, solid bar in Figure~\ref{fig:teaser}).
A benchmark released when RoboDojo saturates, in \forecastNinetyDate{}, would instead need tasks from about \dojoNinety{} to \needHi{}~Elo points to last another year at the pace since G0.5 (the desired benchmark, dashed bar); on the trend, the best policy would already score about \nextTrendScore{} on its median task at release and saturate it within about \nextTrendMonths{} months.

\paragraph{What remains hard are open-ended tasks and randomized scenes.}
Figure~\ref{fig:usedup}b groups the tasks by category: RoboDojo's five capability dimensions and LIBERO's four suites; RoboTwin~2.0 and SimplerEnv have no official categories, so their tasks are grouped by condition.
RoboDojo's Open tasks, which are open-ended or conditioned on language or images, are the hardest category (median 2782~Elo points): today's best policy lies below seven of the eight, and five are already as hard as the tasks a next benchmark needs.
Memory, Precision and Generalization tasks lie around today's best level, while six of the eight Long-Horizon tasks lie below it.
Randomizing the scene makes every task harder, in all 69 pairs of the same task with and without randomization: by a median of 424~Elo points on RoboTwin~2.0, which makes a task about 11 times harder and lowers an expected score of 50\% to 8\%, by 256 on RoboDojo (about 4 times; 50\% to 19\%) and by 77 on SimplerEnv (50\% to 39\%).
A next benchmark that is meant to last should therefore build on open-ended tasks in randomized scenes.

\section{Methodology}
\label{sec:method}

\begin{figure*}[ht]
  \centering
  \includegraphics[width=\linewidth]{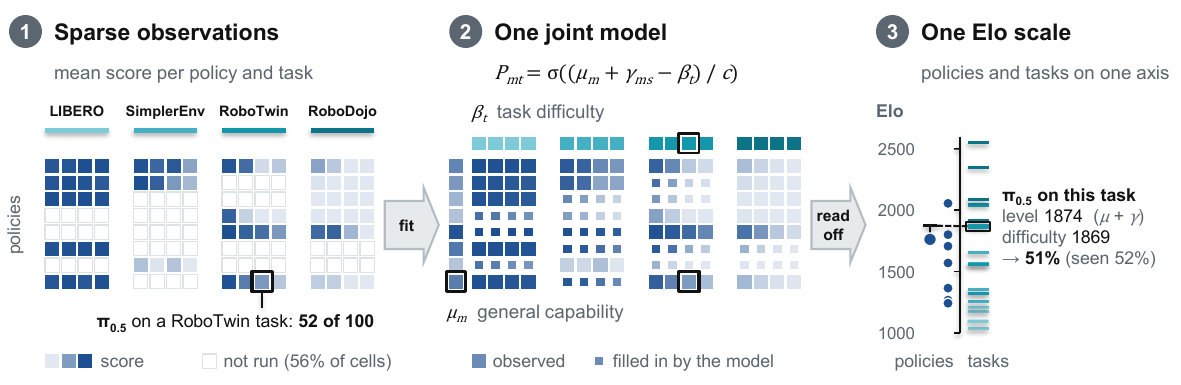}
  \caption{\textbf{From sparse per-task results to one Elo scale.}
  (1)~Observed scores of eight policies on four tasks per benchmark. (2)~One model predicts every cell from policy capabilities and task difficulties; small squares: cells filled in by the model. (3)~Both lie on one Elo scale. Outlined: one example cell.}
  \label{fig:method}
\end{figure*}

Our framework turns per-task results from several benchmarks into one scale (Figure~\ref{fig:method}).

\subsection{Data Representation}
\label{sec:method-data}

The basic unit of data is a \emph{cell}: policy $m$ attempted test case $t$ a number of times $n_{mt}$, and each trial earned a score between 0 and 1; $k_{mt}$ is the sum of these scores.
Each cell enters the fit through these two numbers.
We keep the number of trials, not only the mean score, because a mean over more trials is more precise: 75 points in 150 trials and 25 points in 50 trials both give 50\%, but the first holds three times as much evidence, and the likelihood (Section~\ref{sec:method-fit}) gives it three times the weight.

We adopt a partial-success scoring scheme to capture incremental progress.
Even when a trial fails overall, it may complete intermediate steps (e.g., grasping an object without placing it); accordingly, we assign a score of 1 to successful trials and a fractional score to failures based on the proportion of the task completed.
RoboDojo reports such a score for every test case.
In LIBERO, a task's goal is a list of conditions, and a failure scores the share of them that hold at the end.
Most SimplerEnv tasks have two stages, such as grasping a can and then lifting it, and a failure that completes the first stage scores 0.5; on the drawer tasks, it scores the share of the required drawer travel.
RoboTwin~2.0 reports only success, so its trials score 0 or 1.
Appendix~\ref{app:data} details the full set of rules used in this study.

A \emph{test case} is the smallest unit that gets its own difficulty in the model: a task under given conditions.
When a change of conditions makes a task clearly harder or easier, it becomes a separate test case.
For example, SimplerEnv's task of picking up a coke can gives three test cases, because a can laid horizontally, laid vertically or standing upright differs in difficulty; the clean and randomized versions of a RoboTwin~2.0 task are two test cases.
In total the four benchmarks have \nTests{} test cases (Table~\ref{tab:data}).
Test cases are grouped into \emph{settings} $s$, and each policy gets one offset per setting (Section~\ref{sec:method-model}): RoboTwin~2.0 has a clean and a randomized setting, SimplerEnv has a Google Robot and a WidowX setting, and LIBERO and RoboDojo have one setting each.
The three coke-can test cases thus share one setting, while the two versions of a RoboTwin~2.0 task fall into different settings.

Benchmarks are linked through the policies they share: a policy evaluated on two benchmarks ties their scales together.
Most benchmarks expect a policy to be fine-tuned on their own data, so we treat the checkpoints of one model family, fine-tuned for different benchmarks, as one policy, with fine-tuned variants on each setting.
These checkpoints start from the same base model, so we assume that they share the same underlying capability.
Without this assumption, policies fine-tuned for different benchmarks could not be compared at all.

\subsection{Measurement Model}
\label{sec:method-model}

The expected score $P_{mt}$ of policy $m$ on test case $t$ is modelled as
\begin{equation}
  \log \frac{P_{mt}}{1 - P_{mt}} \;=\; \frac{\mu_m + \gamma_{ms} - \beta_t}{c},
  \label{eq:model}
\end{equation}
where $s$ is the setting of test case $t$ and $c = 400/\ln 10 \approx 173.7$ converts Elo points to logits.
Fractional outcomes contribute partial credit, as draws do in chess; on RoboTwin~2.0, where trials score 0 or 1, $P_{mt}$ is the probability of success.
Here $\mu_m$ is the \textbf{general capability} of policy $m$, the part of its level shared by all benchmarks, by which we rank policies; $\gamma_{ms}$ is its \textbf{offset} on setting $s$, so that its \textbf{level} there is $\mu_m + \gamma_{ms}$; and $\beta_t$ is the \textbf{difficulty} of test case $t$, the level at which the expected score is 50\%.

\paragraph{Capability and difficulty.}
The capability--difficulty formulation, shared with Elo ratings, the Bradley--Terry model and the Rasch model \citep{elo1978rating,bradley1952rank,rasch1960probabilistic}, puts setting-specific policy capabilities and test-case difficulties on the same logarithmic scale axis.
Writing $\theta_m = 10^{\mu_m/400}$ for the capability of policy $m$ and $\delta_t = 10^{\beta_t/400}$ for the difficulty of test case $t$, Equation~\eqref{eq:model} becomes
\begin{equation}
  \frac{P_{mt}}{1 - P_{mt}} \;=\; 10^{\gamma_{ms}/400}\,\frac{\theta_m}{\delta_t},
  \label{eq:odds}
\end{equation}
so the odds of success are the ratio of capability to difficulty, adjusted by the policy's offset on the setting.
When we say that one policy is $k$ times as capable as another, we mean that its $\theta$ is $k$ times larger: on any fixed task its odds of success are $k$ times higher (for partial scores, the odds of the expected score).
A gain of $400 \log_{10} 2 \approx 120$~Elo points doubles capability, and a policy 400~Elo points above a task has odds of 10 to 1, an expected score of 91\%, on every benchmark, as in chess.
These ratios do not depend on where the scale is anchored (Section~\ref{sec:method-scale}), because shifting all ratings multiplies every $\theta$ and $\delta$ by the same factor.
Doubling capability does not double the success rate: it lifts an expected score of 10\% to 18\%, 50\% to 67\% and 90\% to 95\%.
Unlike Elo scores, which are updated game by game, the ratings are parameters of one model fitted to all results at once, so they do not depend on the order in which results arrive.

\paragraph{Setting-specific offsets.}
Robot policies are often fine-tuned separately for each benchmark, so their capabilities need not transfer uniformly across benchmarks.
OpenVLA-OFT~\citep{kim2025openvlaoft}, for example, succeeds on 97\% of LIBERO trials, yet its level on LIBERO lies 657~Elo points above its level on RoboDojo, and by general capability it ranks only 38th of \nPolicies{}.
Task difficulties capture challenges shared across policies, whereas the offset $\gamma_{ms}$ captures policy-specific departures from general capability on setting $s$.
In IRT, this is a testlet effect \citep{bradlow1999bayesian}.
We give each setting, not just each benchmark, its own offset, because a change of conditions within a benchmark, such as randomized scenes or another robot, can help or hurt different policies by very different amounts, which one offset per benchmark cannot express.

\paragraph{Common unit across benchmarks.}
The model is a Rasch testlet model: it has no discrimination parameter, the factor $a$ of the 2-parameter logistic model \citep{birnbaum1968some} that would let the same gap in rating change the score more on some benchmarks or tasks than on others.
We tested such factors, and they do not predict better.
One factor per benchmark or per setting changes the held-out log loss of random-cell cross-validation by at most 0.0001 (\cvLossShared{} with one shared unit).
In the hidden-benchmark test (Appendix~\ref{app:checks}), one factor per benchmark predicts slightly worse (\linkLossPerBench{} against \linkLossShared{}) and one per setting slightly better (\linkLossPerSetting{}).
One factor per test case predicts worse in both tests.
Without such factors, a gap in Elo has the same meaning on every benchmark, and the prior widths below hold on the reported scale.

\subsection{Priors and Estimation}
\label{sec:method-fit}

Maximum likelihood alone breaks down in three places.
A task on which every trial scored 0 gets an infinite difficulty; a policy whose trials all scored 0 gets a capability of minus infinity; and for a policy evaluated on only one benchmark, the data determine only the sum $\mu_m + \gamma_{ms}$, not its two parts.
Weak priors keep every estimate finite and let similar quantities borrow strength from each other.
We state the prior widths in logits; one logit equals $c \approx 174$~Elo points.
\begin{itemize}
  \item General capability: $\mu_m \sim \mathcal{N}(m_0, \sigma_\mu^2)$ with $\sigma_\mu = 2$ logits; $m_0$ is estimated.
  \item Offsets: $\gamma_{ms} \sim \mathcal{N}(0, \sigma_\gamma^2)$ with $\sigma_\gamma = 1$ logit.
  \item Difficulties: $\beta_t \sim \mathcal{N}(\eta_g, \sigma_\beta^2)$ with $\sigma_\beta = 8$ logits, where $\eta_g$ is the mean difficulty of the task's group $g$ (for example, one LIBERO suite) and $\eta_g \sim \mathcal{N}(0, 3^2)$.
\end{itemize}

The width $\sigma_\gamma$ is the most important choice: a small $\sigma_\gamma$ lets a policy's results on one benchmark say a lot about the others, a large one treats every benchmark as separate, and it also decides how a one-benchmark policy's level is split between $\mu_m$ and $\gamma_{ms}$.
We chose $\sigma_\gamma$ and $\sigma_\beta$ with the hidden-benchmark test of Appendix~\ref{sec:linking}.

We fit all cells at once by maximising the log-likelihood $\sum_{mt} \bigl[k_{mt} \log P_{mt} + (n_{mt} - k_{mt}) \log (1 - P_{mt})\bigr]$ plus the log-priors (a maximum a posteriori estimate), using L-BFGS-B with analytic gradients.
When every trial scores 0 or 1 this is the binomial likelihood; with partial scores it is the quasi-likelihood of fractional responses \citep{papke1996econometric}, which still estimates the expected score correctly.
A fit takes a few seconds on one CPU core.
95\% confidence intervals (CIs) come from 1,000 bootstrap refits, each of which resamples the test cases within every setting; for the slopes of policy trends over time, the policies are resampled as well, unless stated otherwise.
We resample test cases because a rating also depends on which tasks a benchmark happens to contain, an uncertainty that the likelihood alone ignores; resampling them also absorbs part of the extra spread between cells (Appendix~\ref{app:checks}).

\subsection{Calibration and Anchoring}
\label{sec:method-scale}

Two changes leave every predicted score unchanged, so the data cannot decide them:
(i) adding the same number to all offsets and all difficulties of one setting; and
(ii) shifting all ratings together.
We fix them by setting (i) the mean offset of a fixed \emph{reference panel} of policies to zero on each setting and (ii) the mean general capability of the reference panel to 1500.
We apply both conventions after the fit to label the scale while preserving every prediction.

The reference panel consists of seven open policies that we can run ourselves and that span a wide range of capability: \pizero{}~\citep{black2024pi0}, \pihalf{}, GR00T~N1.7~\citep{nvidia2025gr00tn1}, SmolVLA~\citep{shukor2025smolvla}, OpenVLA-OFT, X-VLA~\citep{zheng2025xvla} and ACT~\citep{zhao2023act}.
A reference policy counts on a setting only if its scores there are neither all 0 nor all 1; Appendix~\ref{app:anchors} gives the rule and the anchors used where too few qualify.

Convention~(i) is the assumption on which every comparison across benchmarks rests: the reference policies, on average, perform on every benchmark as their general capability predicts.
Because it changes no prediction, no data can confirm it; what can be tested is a necessary condition, that a policy's results on some benchmarks predict its results on another (Appendix~\ref{sec:linking}).
The ranking by general capability does not depend on which policies form the reference panel: changing the panel shifts every $\mu_m$ by the same amount.

\paragraph{Ranking criterion.}
We rank policies by $\mu_m$ rather than by their average level over the benchmarks they ran.
The average level would reward a policy for running only its best benchmark.
For a policy seen on only one benchmark, $\mu_m$ instead keeps part of the policy's distance from the average policy (\shrinkShare{} with our data) and assigns the rest to the offset.
A policy's level on each setting, $\mu_m + \gamma_{ms}$, remains available whenever a benchmark-specific comparison is needed.
We therefore use general capability for the ranking and for the growth of the field, and levels for questions about one benchmark, such as how much of it is left and when it will saturate (Section~\ref{sec:left}).

\section{Related Work}
\label{sec:related}

Placing test takers and test items on one scale is the core of item response theory (IRT) \citep{rasch1960probabilistic,birnbaum1968some}, which shares its logistic form with Elo and Bradley--Terry ratings \citep{elo1978rating,bradley1952rank}.
Educational testing links different test forms through the examinees they share, preferably by fitting all forms jointly \citep{kolen2014test,hanson2002obtaining}, and vertical scaling uses such links to follow students' growth across school grades \citep{tong2007comparisons}.
AI evaluation has adopted these models to analyse test items, shrink benchmarks, recover rankings from sparse results and rate models together with test cases \citep{martinezplumed2019item,rodriguez2021evaluation,polo2024tinybenchmarks,kipnis2025metabench,kang2026scaling,sun2025agielo}, and Chatbot Arena rates language models from pairwise human votes \citep{chiang2024chatbot}.
These methods work within one benchmark or a fixed set of benchmarks.
Our framework uses test linking to join benchmarks that different policies have run, with testlet-style offsets \citep{bradlow1999bayesian} for benchmark-specific fine-tuning.

Benchmarks saturate quickly, often soon after release \citep{ott2022mapping}, and one response is to keep building harder tests \citep{kiela2021dynabench}.
To follow progress across saturated benchmarks, observational scaling laws extract a few capability measures from many language-model benchmarks \citep{ruan2024observational}, METR measures the length of the tasks that models can complete \citep{kwa2025measuring}, and the Rosetta Stone fits one capability per model and one difficulty per benchmark to aggregate scores in order to forecast progress \citep{ho2025rosetta}.
These works study language models and use benchmark averages.
Our framework works with per-task results, so it also gives every task a difficulty, and it provides the first measurement of the pace of progress in robot manipulation.

Robot manipulation policies are compared mainly on simulation benchmarks such as LIBERO, SimplerEnv, RoboTwin~2.0 and RoboDojo \citep{liu2023libero,li2024simpler,chen2025robotwin2,chen2026robodojo}, and perturbed versions such as LIBERO-Plus show how fragile near-perfect scores can be \citep{fei2025liberoplus}.
Real-world evaluation has moved to shared platforms that run new trials, with automatic resets \citep{zhou2025autoeval}, crowd-sourced pairwise comparisons \citep{atreya2025roboarena}, or more than a hundred sites, where rankings agree only weakly across robot embodiments \citep{duan2026robocolosseum}.
Other work evaluates policies in reconstructed scenes or learned world models \citep{jangir2026robotarena,li2025worldeval,quevedo2025worldgym}, or improves the statistics of policy comparisons \citep{kressgazit2024robot,snyder2025your}.
All of these compare policies within one benchmark or platform and need new trials.
Our framework reuses existing results and places the benchmarks on one scale.

\section{Conclusion and Future Work}
\label{sec:conclusion}

We presented a statistical framework that places policies and tasks from different embodied AI benchmarks on one Elo scale, using per-task results that already exist.
Policies evaluated on several benchmarks link them, so two policies can be compared even when they never ran the same benchmark, and the scale predicts results on benchmarks hidden from the fit.
On this scale, robot manipulation policies improve quickly, and since June 2026 faster than new benchmarks grow harder; every benchmark except RoboDojo is already saturated.
If the pace since June 2026 holds, RoboDojo will saturate around \forecastNinety{}, so the field will soon need a harder benchmark, and our results suggest building it on open-ended tasks in randomized scenes.

\paragraph{Future work.}
First, more policies and benchmarks can join the scale, widening the range of tasks, robots and policy types; and the Claude models in Figure~\ref{fig:teaser} can join the fit once their per-task results on RoboDojo are published.
Second, real-robot benchmarks with fixed tasks, such as the real-world track of RoboDojo \citep{chen2026robodojo}, can join through the policies they share with simulation, which would test whether progress in simulation carries over to real robots.
Finally, refitting the scale as new results appear turns the forecast into a test: whether RoboDojo saturates around \forecastNinety{} will show whether the pace since June 2026 lasts.

\bibliography{refs}

\begin{thebibliography}{85}
\providecommand{\natexlab}[1]{#1}
\providecommand{\url}[1]{\texttt{#1}}
\expandafter\ifx\csname urlstyle\endcsname\relax
  \providecommand{\doi}[1]{doi: #1}\else
  \providecommand{\doi}{doi: \begingroup \urlstyle{rm}\Url}\fi

\bibitem[{\'A}fra(2024)]{afra2024oidn}
{\'A}fra, A.~T.
\newblock {Intel Open Image Denoise}.
\newblock \url{https://www.openimagedenoise.org}, 2024.

\bibitem[{Anthropic}(2026)]{anthropic2026fable52}
{Anthropic}.
\newblock Introducing {Claude Fable 5.2} and {Claude Mythos 5.2}.
\newblock \url{https://www.anthropic.com/news/claude-fable-and-mythos-5-2},
  November 2026.

\bibitem[Atreya et~al.(2025)Atreya, Pertsch, Lee, et~al.]{atreya2025roboarena}
Atreya, P., Pertsch, K., Lee, T., et~al.
\newblock {RoboArena}: Distributed real-world evaluation of generalist robot
  policies.
\newblock arXiv:2506.18123, 2025.

\bibitem[Birnbaum(1968)]{birnbaum1968some}
Birnbaum, A.
\newblock Some latent trait models and their use in inferring an examinee's
  ability.
\newblock In Lord, F.~M. and Novick, M.~R. (eds.), \emph{Statistical Theories
  of Mental Test Scores}. Addison-Wesley, Reading, MA, 1968.

\bibitem[Black et~al.(2024)]{black2024pi0}
Black, K. et~al.
\newblock $\pi_0$: A vision-language-action flow model for general robot
  control.
\newblock arXiv:2410.24164, 2024.

\bibitem[Bradley \& Terry(1952)Bradley and Terry]{bradley1952rank}
Bradley, R.~A. and Terry, M.~E.
\newblock Rank analysis of incomplete block designs: {I}. the method of paired
  comparisons.
\newblock \emph{Biometrika}, 39\penalty0 (3/4):\penalty0 324--345, 1952.

\bibitem[Bradlow et~al.(1999)Bradlow, Wainer, and Wang]{bradlow1999bayesian}
Bradlow, E.~T., Wainer, H., and Wang, X.
\newblock A {B}ayesian random effects model for testlets.
\newblock \emph{Psychometrika}, 64\penalty0 (2):\penalty0 153--168, 1999.

\bibitem[Briggs \& Weeks(2009)Briggs and Weeks]{briggs2009impact}
Briggs, D.~C. and Weeks, J.~P.
\newblock The impact of vertical scaling decisions on growth interpretations.
\newblock \emph{Educational Measurement: Issues and Practice}, 28\penalty0
  (4):\penalty0 3--14, 2009.

\bibitem[Brohan et~al.(2022)]{brohan2022rt1}
Brohan, A. et~al.
\newblock {RT-1}: Robotics transformer for real-world control at scale.
\newblock arXiv:2212.06817, 2022.

\bibitem[Bu et~al.(2025)]{bu2025univla}
Bu, Q. et~al.
\newblock {UniVLA}: Learning to act anywhere with task-centric latent actions.
\newblock arXiv:2505.06111, 2025.

\bibitem[Cai et~al.(2026)]{cai2026xiaomir0}
Cai, R. et~al.
\newblock {Xiaomi-Robotics-0}: An open-sourced vision-language-action model
  with real-time execution.
\newblock arXiv:2602.12684, 2026.

\bibitem[Chen et~al.(2026{\natexlab{a}})]{chen2025robotwin2}
Chen, T. et~al.
\newblock {RoboTwin 2.0}: A scalable data generator and benchmark with strong
  domain randomization for robust bimanual robotic manipulation.
\newblock In \emph{Proceedings of the 43rd International Conference on Machine
  Learning (ICML)}, 2026{\natexlab{a}}.
\newblock arXiv:2506.18088.

\bibitem[Chen et~al.(2026{\natexlab{b}})]{chen2026robodojo}
Chen, T. et~al.
\newblock {RoboDojo}: A unified sim-and-real benchmark for comprehensive
  evaluation of generalist robot manipulation policies.
\newblock arXiv:2607.04434, 2026{\natexlab{b}}.

\bibitem[Chen et~al.(2025)]{chen2025internvlam1}
Chen, X. et~al.
\newblock {InternVLA-M1}: A spatially guided vision-language-action framework
  for generalist robot policy.
\newblock arXiv:2510.13778, 2025.

\bibitem[Chi et~al.(2023)]{chi2023diffusion}
Chi, C. et~al.
\newblock {Diffusion Policy}: Visuomotor policy learning via action diffusion.
\newblock arXiv:2303.04137, 2023.

\bibitem[Chiang et~al.(2024)Chiang, Zheng, Sheng, Angelopoulos, Li, Li, Zhu,
  Zhang, Jordan, Gonzalez, and Stoica]{chiang2024chatbot}
Chiang, W.-L., Zheng, L., Sheng, Y., Angelopoulos, A.~N., Li, T., Li, D., Zhu,
  B., Zhang, H., Jordan, M.~I., Gonzalez, J.~E., and Stoica, I.
\newblock Chatbot arena: An open platform for evaluating {LLMs} by human
  preference.
\newblock In \emph{Proceedings of the 41st International Conference on Machine
  Learning (ICML)}, 2024.

\bibitem[{DeepSeek}(2026)]{deepseek2026v41flash}
{DeepSeek}.
\newblock {DeepSeek-V4.1-Flash}: Smarter, faster, more efficient.
\newblock \url{https://api-docs.deepseek.com/news/news260910}, 2026.

\bibitem[{Dexmal}(2026)]{dexmal2026dm05}
{Dexmal}.
\newblock {DM0.5}: Designed for the open world, where generalization emerges.
\newblock \url{https://www.dexmal.com/blog/dm0.5/index_en.html}, 2026.

\bibitem[Elo(1978)]{elo1978rating}
Elo, A.~E.
\newblock \emph{The Rating of Chessplayers, Past and Present}.
\newblock Arco Publishing, New York, 1978.

\bibitem[Fang et~al.(2026)]{fang2026molmoact2}
Fang, H. et~al.
\newblock {MolmoAct2}: Action reasoning models for real-world deployment.
\newblock arXiv:2605.02881, 2026.

\bibitem[Fei et~al.(2025)Fei, Wang, et~al.]{fei2025liberoplus}
Fei, S., Wang, S., et~al.
\newblock {LIBERO-Plus}: In-depth robustness analysis of vision-language-action
  models.
\newblock arXiv:2510.13626, 2025.

\bibitem[{Galaxea}(2026)]{galaxea2026g05}
{Galaxea}.
\newblock {Galaxea G0.5}: One autoregressive stream for robot reasoning and
  action.
\newblock \url{https://opengalaxea.github.io/G05/}, 2026.

\bibitem[{GigaBrain Team} et~al.(2026)]{gigabrain2026gigabrain07}
{GigaBrain Team} et~al.
\newblock {GigaBrain-0.7}: Scaling embodied foundation models to emergent
  capabilities with a three-system architecture.
\newblock arXiv:2608.15875, 2026.

\bibitem[Guo et~al.(2026)]{guo2026vpp2}
Guo, Y. et~al.
\newblock {Video Prediction Policy 2}: Predict better, act better.
\newblock arXiv:2610.10270, 2026.

\bibitem[Hanson \& B{\'e}guin(2002)Hanson and B{\'e}guin]{hanson2002obtaining}
Hanson, B.~A. and B{\'e}guin, A.~A.
\newblock Obtaining a common scale for item response theory item parameters
  using separate versus concurrent estimation in the common-item equating
  design.
\newblock \emph{Applied Psychological Measurement}, 26\penalty0 (1):\penalty0
  3--24, 2002.

\bibitem[Ho et~al.(2025)Ho, Denain, Atanasov, Albanie, and Shah]{ho2025rosetta}
Ho, A., Denain, J.-S., Atanasov, D., Albanie, S., and Shah, R.
\newblock A {R}osetta {S}tone for {AI} benchmarks.
\newblock arXiv:2512.00193, 2025.

\bibitem[Jangir et~al.(2026)Jangir, Zhang, et~al.]{jangir2026robotarena}
Jangir, Y., Zhang, Y., et~al.
\newblock {RobotArena} $\infty$: Scalable robot benchmarking via real-to-sim
  translation.
\newblock In \emph{International Conference on Learning Representations
  (ICLR)}, 2026.

\bibitem[Jiang et~al.(2026)Jiang, Tan, Wheeler, Sun, Ayalew, and
  Walter]{jiang2026benchmarking}
Jiang, T., Tan, X., Wheeler, S., Sun, L., Ayalew, T.~W., and Walter, M.
\newblock What are we actually benchmarking in robot manipulation?
\newblock arXiv:2606.04233, 2026.

\bibitem[Kang(2026)]{kang2026scaling}
Kang, J.~M.
\newblock The scaling law of evaluation failure: Why simple averaging collapses
  under data sparsity and item difficulty gaps, and how item response theory
  recovers ground truth across domains.
\newblock arXiv:2605.11205, 2026.

\bibitem[Kiela et~al.(2021)]{kiela2021dynabench}
Kiela, D. et~al.
\newblock Dynabench: Rethinking benchmarking in {NLP}.
\newblock In \emph{Proceedings of the 2021 Conference of the North American
  Chapter of the Association for Computational Linguistics (NAACL)}, pp.\
  4110--4124, 2021.

\bibitem[Kim et~al.(2025)Kim, Finn, and Liang]{kim2025openvlaoft}
Kim, M.~J., Finn, C., and Liang, P.
\newblock Fine-tuning vision-language-action models: Optimizing speed and
  success.
\newblock arXiv:2502.19645, 2025.

\bibitem[Kim et~al.(2024)]{kim2024openvla}
Kim, M.~J. et~al.
\newblock {OpenVLA}: An open-source vision-language-action model.
\newblock arXiv:2406.09246, 2024.

\bibitem[Kipnis et~al.(2025)Kipnis, Voudouris, Schulze~Buschoff, and
  Schulz]{kipnis2025metabench}
Kipnis, A., Voudouris, K., Schulze~Buschoff, L.~M., and Schulz, E.
\newblock metabench -- a sparse benchmark of reasoning and knowledge in large
  language models.
\newblock In \emph{International Conference on Learning Representations
  (ICLR)}, 2025.

\bibitem[Kolen \& Brennan(2014)Kolen and Brennan]{kolen2014test}
Kolen, M.~J. and Brennan, R.~L.
\newblock \emph{Test Equating, Scaling, and Linking: Methods and Practices}.
\newblock Springer, New York, 3rd edition, 2014.

\bibitem[Kress-Gazit et~al.(2024)]{kressgazit2024robot}
Kress-Gazit, H. et~al.
\newblock Robot learning as an empirical science: Best practices for policy
  evaluation.
\newblock arXiv:2409.09491, 2024.

\bibitem[Kwa et~al.(2025)]{kwa2025measuring}
Kwa, T. et~al.
\newblock Measuring {AI} ability to complete long tasks.
\newblock arXiv:2503.14499, 2025.

\bibitem[Lee et~al.(2025)]{lee2025molmoact}
Lee, J. et~al.
\newblock {MolmoAct}: Action reasoning models that can reason in space.
\newblock arXiv:2508.07917, 2025.

\bibitem[Li et~al.(2026)]{li2026lingbotva}
Li, L. et~al.
\newblock Causal world modeling for robot control.
\newblock arXiv:2601.21998, 2026.

\bibitem[Li et~al.(2024{\natexlab{a}})]{li2024cogact}
Li, Q. et~al.
\newblock {CogACT}: A foundational vision-language-action model for synergizing
  cognition and action in robotic manipulation.
\newblock arXiv:2411.19650, 2024{\natexlab{a}}.

\bibitem[Li et~al.(2024{\natexlab{b}})]{li2024simpler}
Li, X. et~al.
\newblock Evaluating real-world robot manipulation policies in simulation.
\newblock In \emph{Proceedings of the 8th Conference on Robot Learning (CoRL)},
  2024{\natexlab{b}}.

\bibitem[Li et~al.(2025)Li, Zhu, Wen, Shen, and Xu]{li2025worldeval}
Li, Y., Zhu, Y., Wen, J., Shen, C., and Xu, Y.
\newblock {WorldEval}: World model as real-world robot policies evaluator.
\newblock arXiv:2505.19017, 2025.

\bibitem[Liu et~al.(2023)Liu, Zhu, Gao, Feng, Liu, Zhu, and
  Stone]{liu2023libero}
Liu, B., Zhu, Y., Gao, C., Feng, Y., Liu, Q., Zhu, Y., and Stone, P.
\newblock {LIBERO}: Benchmarking knowledge transfer for lifelong robot
  learning.
\newblock In \emph{Advances in Neural Information Processing Systems (NeurIPS),
  Datasets and Benchmarks Track}, 2023.

\bibitem[Liu et~al.(2024)]{liu2024rdt}
Liu, S. et~al.
\newblock {RDT-1B}: a diffusion foundation model for bimanual manipulation.
\newblock arXiv:2410.07864, 2024.

\bibitem[Ma et~al.(2026)]{ma2026internvlaa15}
Ma, H. et~al.
\newblock {InternVLA-A1.5}: Unifying understanding, latent foresight, and
  action for compositional generalization.
\newblock arXiv:2607.04988, 2026.

\bibitem[Maia~Polo et~al.(2024)]{polo2024tinybenchmarks}
Maia~Polo, F. et~al.
\newblock tiny{B}enchmarks: Evaluating {LLMs} with fewer examples.
\newblock In \emph{Proceedings of the 41st International Conference on Machine
  Learning (ICML)}, pp.\  34303--34326, 2024.

\bibitem[Mart{\'i}nez-Plumed et~al.(2019)Mart{\'i}nez-Plumed, Prud{\^e}ncio,
  Mart{\'i}nez-Us{\'o}, and Hern{\'a}ndez-Orallo]{martinezplumed2019item}
Mart{\'i}nez-Plumed, F., Prud{\^e}ncio, R. B.~C., Mart{\'i}nez-Us{\'o}, A., and
  Hern{\'a}ndez-Orallo, J.
\newblock Item response theory in {AI}: Analysing machine learning classifiers
  at the instance level.
\newblock \emph{Artificial Intelligence}, 2019.

\bibitem[Miao et~al.(2026)]{miao2026internw0}
Miao, X. et~al.
\newblock {InternW0-$\Delta$}: A world action model bridging predictive
  dynamics and actions with {20K+} hours of open data.
\newblock arXiv:2609.31394, 2026.

\bibitem[Nguyen et~al.(2026)Nguyen, Wang, Shailesh, Yang, Hu, Jiang, Fang, Ren,
  Luo, Chong, Tung, Pravisani, Lin, Mo, Singh, Tee, Soh, Srinivasa, Fox, Hsu,
  Song, Ellerbach, Wolf, Kooijmans, He, Cho, and Duan]{duan2026robocolosseum}
Nguyen, T.~S., Wang, Y.~R., Shailesh, S., Yang, G., Hu, J., Jiang, J., Fang,
  H., Ren, Z., Luo, Z., Chong, I., Tung, A., Pravisani, S., Lin, M., Mo, Y.,
  Singh, I., Tee, B., Soh, H., Srinivasa, S., Fox, D., Hsu, D., Song, H.,
  Ellerbach, M., Wolf, T., Kooijmans, P., He, R., Cho, M., and Duan, J.
\newblock {RoboColosseum}: Evaluating physical intelligence at scale
  in-the-wild.
\newblock arXiv:2611.02497, 2026.

\bibitem[{NVIDIA}(2025)]{nvidia2025gr00tn15simpler}
{NVIDIA}.
\newblock {Isaac GR00T} {N1.5}: Fine-tuning and evaluation on {SimplerEnv}.
\newblock
  \url{https://github.com/NVIDIA/Isaac-GR00T/tree/n1.5-release/examples/SimplerEnv},
  2025.

\bibitem[{NVIDIA} et~al.(2025){NVIDIA}, Bjorck, et~al.]{nvidia2025gr00tn1}
{NVIDIA}, Bjorck, J., et~al.
\newblock {GR00T N1}: An open foundation model for generalist humanoid robots.
\newblock arXiv:2503.14734, 2025.

\bibitem[{Octo Model Team} et~al.(2024)]{octo2024octo}
{Octo Model Team} et~al.
\newblock {Octo}: An open-source generalist robot policy.
\newblock arXiv:2405.12213, 2024.

\bibitem[{Open X-Embodiment Collaboration} et~al.(2023)]{oxe2023rtx}
{Open X-Embodiment Collaboration} et~al.
\newblock {Open X-Embodiment}: Robotic learning datasets and {RT-X} models.
\newblock arXiv:2310.08864, 2023.

\bibitem[{OpenAI}(2026{\natexlab{a}})]{openai2026gpt55}
{OpenAI}.
\newblock Introducing {GPT-5.5}.
\newblock \url{https://openai.com/index/introducing-gpt-5-5/},
  2026{\natexlab{a}}.

\bibitem[{OpenAI}(2026{\natexlab{b}})]{openai2026gpt6astra}
{OpenAI}.
\newblock {GPT-6 Astra}.
\newblock \url{https://openai.com/index/gpt-6-astra/}, 2026{\natexlab{b}}.

\bibitem[Ott et~al.(2022)Ott, Barbosa-Silva, Blagec, Brauner, and
  Samwald]{ott2022mapping}
Ott, S., Barbosa-Silva, A., Blagec, K., Brauner, J., and Samwald, M.
\newblock Mapping global dynamics of benchmark creation and saturation in
  artificial intelligence.
\newblock \emph{Nature Communications}, 13:\penalty0 6793, 2022.

\bibitem[Papke \& Wooldridge(1996)Papke and Wooldridge]{papke1996econometric}
Papke, L.~E. and Wooldridge, J.~M.
\newblock Econometric methods for fractional response variables with an
  application to 401(k) plan participation rates.
\newblock \emph{Journal of Applied Econometrics}, 11\penalty0 (6):\penalty0
  619--632, 1996.

\bibitem[Parker et~al.(2010)Parker, Bigler, Dietrich, Friedrich, Hoberock,
  Luebke, McAllister, McGuire, Morley, Robison, and Stich]{parker2010optix}
Parker, S.~G., Bigler, J., Dietrich, A., Friedrich, H., Hoberock, J., Luebke,
  D., McAllister, D., McGuire, M., Morley, K., Robison, A., and Stich, M.
\newblock {OptiX}: A general purpose ray tracing engine.
\newblock \emph{ACM Transactions on Graphics}, 29\penalty0 (4):\penalty0
  66:1--66:13, 2010.

\bibitem[Pertsch et~al.(2025)]{pertsch2025fast}
Pertsch, K. et~al.
\newblock {FAST}: Efficient action tokenization for vision-language-action
  models.
\newblock arXiv:2501.09747, 2025.

\bibitem[{Physical Intelligence} et~al.(2025)]{pi2025pi05}
{Physical Intelligence} et~al.
\newblock $\pi_{0.5}$: a vision-language-action model with open-world
  generalization.
\newblock arXiv:2504.16054, 2025.

\bibitem[Qu et~al.(2025)]{qu2025spatialvla}
Qu, D. et~al.
\newblock {SpatialVLA}: Exploring spatial representations for
  visual-language-action model.
\newblock arXiv:2501.15830, 2025.

\bibitem[Quevedo et~al.(2025)]{quevedo2025worldgym}
Quevedo, J. et~al.
\newblock {WorldGym}: World model as an environment for policy evaluation.
\newblock arXiv:2506.00613, 2025.

\bibitem[Rasch(1960)]{rasch1960probabilistic}
Rasch, G.
\newblock \emph{Probabilistic Models for Some Intelligence and Attainment
  Tests}.
\newblock Danish Institute for Educational Research, Copenhagen, 1960.

\bibitem[Rodriguez et~al.(2021)Rodriguez, Barrow, Hoyle, Lalor, Jia, and
  Boyd-Graber]{rodriguez2021evaluation}
Rodriguez, P., Barrow, J., Hoyle, A., Lalor, J.~P., Jia, R., and Boyd-Graber,
  J.
\newblock Evaluation examples are not equally informative: How should that
  change {NLP} leaderboards?
\newblock In \emph{Proceedings of the 59th Annual Meeting of the Association
  for Computational Linguistics (ACL)}, pp.\  4486--4503, 2021.

\bibitem[Ruan et~al.(2024)Ruan, Maddison, and Hashimoto]{ruan2024observational}
Ruan, Y., Maddison, C.~J., and Hashimoto, T.
\newblock Observational scaling laws and the predictability of language model
  performance.
\newblock In \emph{Advances in Neural Information Processing Systems
  (NeurIPS)}, 2024.

\bibitem[Shukor et~al.(2025)]{shukor2025smolvla}
Shukor, M. et~al.
\newblock {SmolVLA}: A vision-language-action model for affordable and
  efficient robotics.
\newblock arXiv:2506.01844, 2025.

\bibitem[Snyder et~al.(2025)]{snyder2025your}
Snyder, D. et~al.
\newblock Is your imitation learning policy better than mine? {P}olicy
  comparison with near-optimal stopping.
\newblock In \emph{Robotics: Science and Systems (RSS)}, 2025.

\bibitem[Sun et~al.(2026)]{sun2026vlajepa}
Sun, J. et~al.
\newblock {VLA-JEPA}: Enhancing vision-language-action model with latent world
  model.
\newblock arXiv:2602.10098, 2026.

\bibitem[Sun et~al.(2025)Sun, Zhao, Lee, Sun, Yuan, Huang, Li, Yeoh, Prakash,
  Malone, and Ang~Jr.]{sun2025agielo}
Sun, S., Zhao, Y., Lee, C. D.~W., Sun, J., Yuan, C., Huang, Z., Li, D., Yeoh,
  J. K.~W., Prakash, A., Malone, T.~W., and Ang~Jr., M.~H.
\newblock {AGI-Elo}: How far are we from mastering a task?
\newblock In \emph{Advances in Neural Information Processing Systems
  (NeurIPS)}, 2025.

\bibitem[Todorov et~al.(2012)Todorov, Erez, and Tassa]{todorov2012mujoco}
Todorov, E., Erez, T., and Tassa, Y.
\newblock {MuJoCo}: A physics engine for model-based control.
\newblock In \emph{IEEE/RSJ International Conference on Intelligent Robots and
  Systems (IROS)}, pp.\  5026--5033, 2012.

\bibitem[Tong \& Kolen(2007)Tong and Kolen]{tong2007comparisons}
Tong, Y. and Kolen, M.~J.
\newblock Comparisons of methodologies and results in vertical scaling for
  educational achievement tests.
\newblock \emph{Applied Measurement in Education}, 20\penalty0 (2):\penalty0
  227--253, 2007.

\bibitem[Wang et~al.(2026{\natexlab{a}})]{wang2026ewam}
Wang, H. et~al.
\newblock {EWAM}: Emergent depth-wise specialization in a unified embodied
  model -- from semantic understanding through visual foresight to action.
\newblock arXiv:2609.39973, 2026{\natexlab{a}}.

\bibitem[Wang et~al.(2026{\natexlab{b}})]{wang2026patchwam}
Wang, T. et~al.
\newblock {An Action Is Worth One Patch}: Unified world-action modeling with
  {PatchWAM}.
\newblock arXiv:2609.25961, 2026{\natexlab{b}}.

\bibitem[{X Square Robot}(2026)]{xsquare2026wallosslite}
{X Square Robot}.
\newblock {WALL-OSS 1.0 Lite -- RoboDojo}.
\newblock
  \url{https://huggingface.co/x-square-robot/wall-oss-1.0-lite-robodojo}, 2026.

\bibitem[{Xiaomi Robotics Team} et~al.(2026)]{xiaomi2026r1}
{Xiaomi Robotics Team} et~al.
\newblock {Xiaomi-Robotics-1}: Scaling vision-language-action models with over
  {100K} hours of real-world trajectories.
\newblock arXiv:2607.15330, 2026.

\bibitem[Xin et~al.(2026)]{xin2026prismwam}
Xin, T. et~al.
\newblock {PrismWAM}: Decoupling state representations in world action models.
\newblock arXiv:2610.26564, 2026.

\bibitem[Yang et~al.(2026)]{yang2026abotm0}
Yang, Y. et~al.
\newblock {ABot-M0}: {VLA Foundation} model for robotic manipulation with
  action manifold learning.
\newblock arXiv:2602.11236, 2026.

\bibitem[Ye et~al.(2026{\natexlab{a}})]{ye2026starvla}
Ye, J. et~al.
\newblock {StarVLA-$\alpha$}: Reducing complexity in vision-language-action
  systems.
\newblock arXiv:2604.11757, 2026{\natexlab{a}}.

\bibitem[Ye et~al.(2026{\natexlab{b}})]{ye2026dreamzero}
Ye, S. et~al.
\newblock World action models are zero-shot policies.
\newblock arXiv:2602.15922, 2026{\natexlab{b}}.

\bibitem[Yuan et~al.(2026)]{yuan2026fastwam}
Yuan, T. et~al.
\newblock {Fast-WAM}: Do world action models need test-time future imagination?
\newblock arXiv:2603.16666, 2026.

\bibitem[Zhang et~al.(2026{\natexlab{a}})]{zhang2026hyvla}
Zhang, H. et~al.
\newblock {Hy-Embodied-0.5-VLA}: From vision-language-action models to a
  real-world robot learning stack.
\newblock arXiv:2606.14409, 2026{\natexlab{a}}.

\bibitem[Zhang et~al.(2026{\natexlab{b}})]{zhang2026unexpected}
Zhang, W. et~al.
\newblock An unexpected robot policy: Early evaluations of {GPT-6 Astra} on
  {RoboDojo} and beyond.
\newblock arXiv:2609.24170, 2026{\natexlab{b}}.

\bibitem[Zhao et~al.(2023)]{zhao2023act}
Zhao, T.~Z. et~al.
\newblock Learning fine-grained bimanual manipulation with low-cost hardware.
\newblock arXiv:2304.13705, 2023.

\bibitem[Zheng et~al.(2025)]{zheng2025xvla}
Zheng, J. et~al.
\newblock {X-VLA}: Soft-prompted transformer as scalable cross-embodiment
  vision-language-action model.
\newblock arXiv:2510.10274, 2025.

\bibitem[Zhou et~al.(2025)Zhou, Atreya, Tan, Pertsch, and
  Levine]{zhou2025autoeval}
Zhou, Z., Atreya, P., Tan, Y.~L., Pertsch, K., and Levine, S.
\newblock {AutoEval}: Autonomous evaluation of generalist robot manipulation
  policies in the real world.
\newblock In \emph{Proceedings of the Conference on Robot Learning (CoRL)},
  2025.

\bibitem[Zitkovich et~al.(2023)]{zitkovich2023rt2}
Zitkovich, B. et~al.
\newblock {RT-2}: Vision-language-action models transfer web knowledge to
  robotic control.
\newblock In \emph{Proceedings of the 7th Conference on Robot Learning (CoRL)},
  2023.
\newblock arXiv:2307.15818.

\end{thebibliography}
\bibliographystyle{icml2027}

\newpage
\appendix
\onecolumn
\renewcommand{\topfraction}{0.9}
\renewcommand{\bottomfraction}{0.8}
\renewcommand{\textfraction}{0.07}
\renewcommand{\floatpagefraction}{0.8}
\setcounter{topnumber}{3}
\setcounter{totalnumber}{4}
\section{Data Collection and Scoring}
\label{app:data}

\begin{figure}[p]
  \centering
  \includegraphics[width=\linewidth,height=0.93\textheight,keepaspectratio]{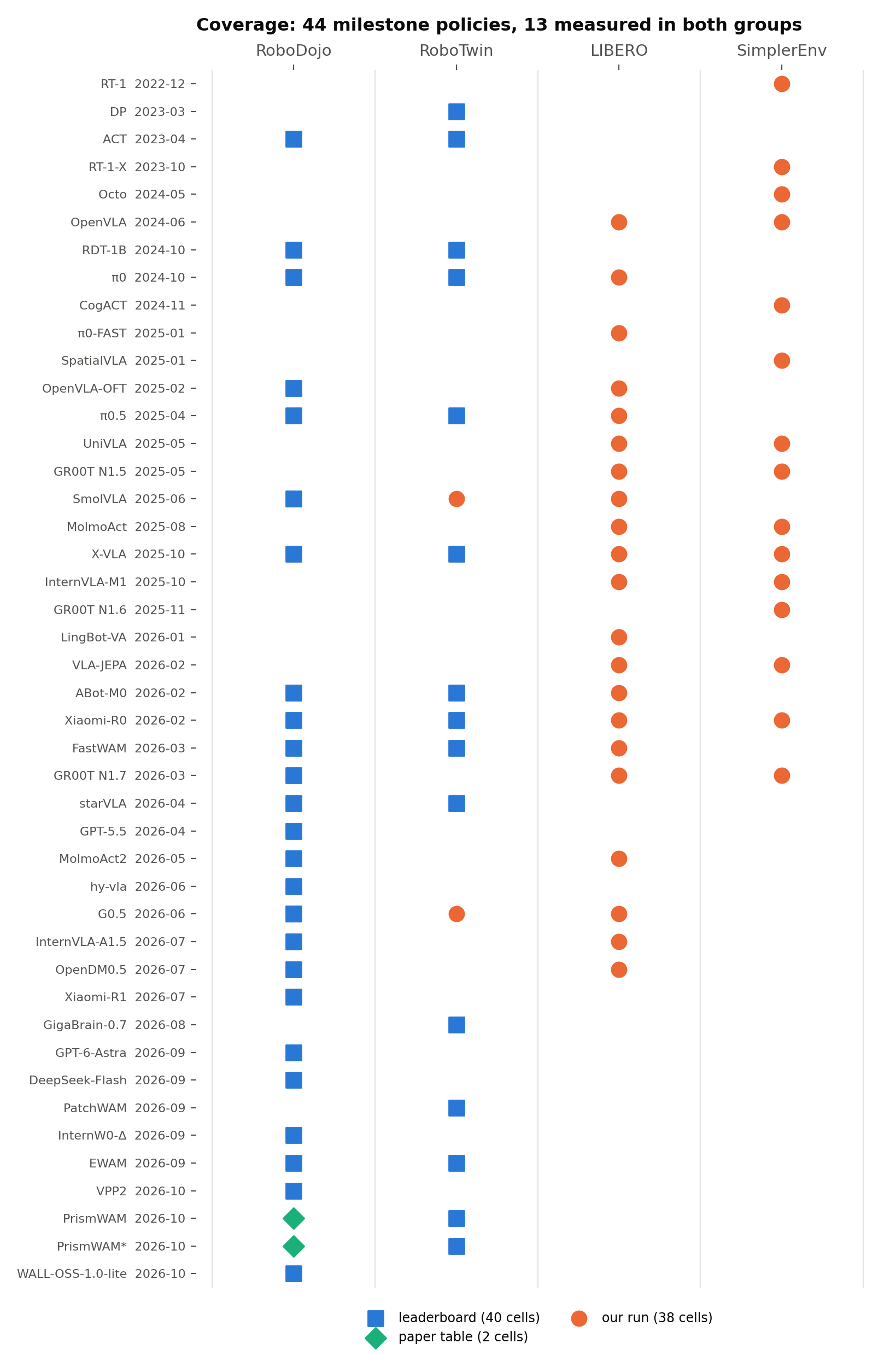}
  \caption{\textbf{Which policy has results on which benchmark, and from which source.} Most cells are empty; \nBothGroups{} policies have results on both a one-arm and a two-arm benchmark.}
  \label{fig:coverage}
\end{figure}

\paragraph{Data sources.}
All data are as of 16 November 2026.
RoboDojo results come from its public leaderboard and, for PrismWAM and PrismWAM$^{*}$, from the per-task table in their paper.
RoboTwin~2.0 results come from its leaderboard.
From RoboDojo's agent track we include the three agents that the RoboDojo team evaluated itself: GPT-6-Astra, GPT-5.5 and DeepSeek-Flash~\citep{deepseek2026v41flash}.
Each is a large language model that the RoboDojo team evaluated zero-shot through the same fixed harness~\citep{zhang2026unexpected}, with no learned policy in the loop, so the three form a controlled comparison of language models as robot agents.
On 14 November 2026 RoboDojo published rules for its agent track and marked entries that deviate from them with an asterisk, listing them unranked; the documented deviations include task-specific solution guides or evaluator tolerances supplied in the prompt.
We date an agent by the public release of its language model (GPT-5.5: April 2026; GPT-6-Astra and DeepSeek-Flash, i.e.\ DeepSeek V4.1-Flash: September 2026).
DeepSeek-Flash was evaluated on 42 of RoboDojo's 54 test cases with 10 or 5 trials each (360 in all), the other two on all 54 with 50 or 25 trials.
We ran LIBERO for 21 policies and SimplerEnv for 15, and RoboTwin~2.0 (clean setting) for SmolVLA and G0.5, which its leaderboard lacks.
Figure~\ref{fig:coverage} shows which policy has results on which benchmark, and from which source.

\paragraph{Evaluation protocols.}
\emph{LIBERO:} four suites of ten tasks; 50 trials per task from the fixed initial states 0--49.
\emph{SimplerEnv:} the official episode lists; Google Robot tasks under both protocols (visual matching and variant aggregation, seven test cases each, with the coke-can task split by the can's orientation) and four WidowX tasks.
\emph{RoboTwin~2.0:} the clean setting of all 50 tasks, 100 trials per task, as on the leaderboard.
We used each policy's own released checkpoint and inference code.
Appendix~\ref{app:settings} shows how much these settings matter.

\paragraph{Partial-success scores.}
A success scores 1; a failure scores as follows, at most 0.99.
\begin{itemize}
  \item \emph{RoboDojo:} the leaderboard's partial-progress score of each test case, divided by 100.
  \item \emph{LIBERO:} the share of the task goal's conditions (BDDL predicates) that hold at the end of the trial. Nine of the 40 tasks have more than one condition; on the others, a failure scores 0.
  \item \emph{SimplerEnv:} 0.5 if the first stage was completed: a stable grasp of the can (pick coke can) or of the object (the four WidowX tasks), moving the right object and no other (move near), or opening the drawer (put the apple into the drawer). On the drawer tasks, the share of the required travel of the drawer: its opening divided by 0.15 (open drawer), or $(0.2 - \text{opening})/0.15$ (close drawer, which starts at an opening of 0.2).
  \item \emph{RoboTwin~2.0:} the leaderboard and our two runs record only success: a failure scores 0.
\end{itemize}
Over all trials, partial credit raises the mean score above the success rate from 10.2\% to 14.3\% on RoboDojo, from 55.0\% to 63.0\% on SimplerEnv and from 91.9\% to 93.1\% on LIBERO.
A task counts as \emph{solved} when some policy succeeded on it at least once.

\paragraph{Trials per cell.}
RoboDojo reports success rates and partial-progress scores averaged over several training seeds, usually three seeds of 50 trials.
We infer the number of seeds from the granularity of each policy's rates (for example, 0.67\% $=$ 1/150) and count all trials.
Seeds are not independent repetitions, so this overstates the precision of RoboDojo cells (Appendix~\ref{app:checks}).

\paragraph{Policy identity across benchmarks.}
A shared name is the only link between two benchmarks, and a wrong link is worse than none: it would move a whole benchmark.
We therefore link two entries only when we can verify that they are the same policy.
Benchmark-specific fine-tuned checkpoints of the same model family (for example, OpenVLA-OFT's LIBERO checkpoint) count as the same policy; the offsets absorb their benchmark-specific capability.

\begin{table}[p]
  \centering
  \caption{\textbf{The \nPolicies{} policies, ranked by general capability $\mu$.} 95\% CI: bootstrap, test cases resampled within each setting. Ran: L~LIBERO, S~SimplerEnv, T~RoboTwin~2.0, D~RoboDojo. $^{\ddagger}$LLM Agent, dated by the release of its language model.}
  \label{tab:policies}
  \small
  \begin{tabular}{@{}rllrll@{}}
    \toprule
    & Policy & Released & $\mu$ & 95\% CI & Ran \\
    \midrule
    \csname @@input\endcsname sections/tab_policies.tex
    \bottomrule
  \end{tabular}
\end{table}

\section{Model Specification}
\label{app:model}

\paragraph{Priors.}
Table~\ref{tab:priors} lists every prior and convention.
The widths $\sigma_\gamma$ and $\sigma_\beta$ were chosen with the hidden-benchmark test (Appendix~\ref{app:checks}).
The model has no conversion factor (discrimination): every benchmark shares the unit $c$.

\begin{table}[htbp]
  \centering
  \caption{\textbf{Priors and conventions.} One logit equals $400/\ln 10 \approx 174$~Elo points.}
  \label{tab:priors}
  \small
  \begin{tabular}{@{}lll@{}}
    \toprule
    Quantity & Prior or convention & Value \\
    \midrule
    General capability $\mu_m$ & $\mathcal{N}(m_0, \sigma_\mu^2)$, $m_0$ estimated & $\sigma_\mu = 2$ logits \\
    Offset $\gamma_{ms}$ & $\mathcal{N}(0, \sigma_\gamma^2)$ & $\sigma_\gamma = 1$ logit \\
    Difficulty $\beta_t$ & $\mathcal{N}(\eta_g, \sigma_\beta^2)$, $\eta_g \sim \mathcal{N}(0, 3^2)$ & $\sigma_\beta = 8$ logits \\
    Shift per setting & mean offset of the anchors $= 0$ & Table~\ref{tab:anchors} \\
    Unit & one logit $= c = 400/\ln 10$ on every benchmark & \\
    Zero point & mean $\mu$ of the reference panel $= 1500$ & \\
    \bottomrule
  \end{tabular}
\end{table}

\paragraph{Anchor policies.}
\label{app:anchors}
On each setting, the offsets and difficulties are shifted so that the mean offset of the \emph{anchors} is zero.
The anchors are the reference policies that are informative on the setting, meaning that their mean score lies strictly between 0 and 1 on at least a quarter of its tasks.
Where fewer than three reference policies qualify, the anchors are the first of these groups with at least three members: informative policies that also ran another benchmark, informative policies, and all policies with data on the setting.
Table~\ref{tab:anchors} lists the result.
Only two reference policies have SimplerEnv results, so SimplerEnv is anchored on other well-measured policies; running more reference policies on SimplerEnv would remove this exception.

\begin{table}[htbp]
  \centering
  \caption{\textbf{Anchors on each setting,} drawn where possible from the reference panel: $\pi_0$, $\pi_{0.5}$, GR00T N1.7, SmolVLA, OpenVLA-OFT, X-VLA and ACT.}
  \label{tab:anchors}
  \small
  \begin{tabular}{@{}lp{9.2cm}@{}}
    \toprule
    Setting & Anchors \\
    \midrule
    RoboDojo & $\pi_0$, $\pi_{0.5}$, GR00T N1.7, SmolVLA, X-VLA, ACT \\
    LIBERO & $\pi_0$, $\pi_{0.5}$, GR00T N1.7, SmolVLA, OpenVLA-OFT, X-VLA \\
    RoboTwin 2.0 clean & $\pi_0$, $\pi_{0.5}$, X-VLA, ACT, SmolVLA \\
    RoboTwin 2.0 randomized & $\pi_0$, $\pi_{0.5}$, X-VLA, ACT \\
    SimplerEnv Google Robot & GR00T N1.7, X-VLA, Xiaomi-R0, GR00T N1.5, InternVLA-M1, MolmoAct, OpenVLA, VLA-JEPA \\
    SimplerEnv WidowX & GR00T N1.7, X-VLA, Xiaomi-R0, GR00T N1.5, InternVLA-M1, OpenVLA, UniVLA, VLA-JEPA \\
    \bottomrule
  \end{tabular}
\end{table}

\paragraph{Alternative summary scores.}
We rank by general capability $\mu_m$.
Two further scores are available for each policy.
The \emph{overall score} is the mean of its levels $\mu_m + \gamma_{ms}$ over the benchmarks it ran, each benchmark counted once.
The \emph{full-coverage score} is the mean over all four benchmarks, with a benchmark the policy did not run filled in by its predicted level, but only where the hidden-benchmark test shows that one of the policy's benchmarks predicts it; 13 policies have such a score.
The overall score depends on which benchmarks a policy happened to run: a policy that also ran a benchmark it does poorly on scores lower than one that did not.
This is why we rank by $\mu$.

\begin{figure}[p]
  \centering
  \includegraphics[width=\linewidth,height=0.93\textheight,keepaspectratio]{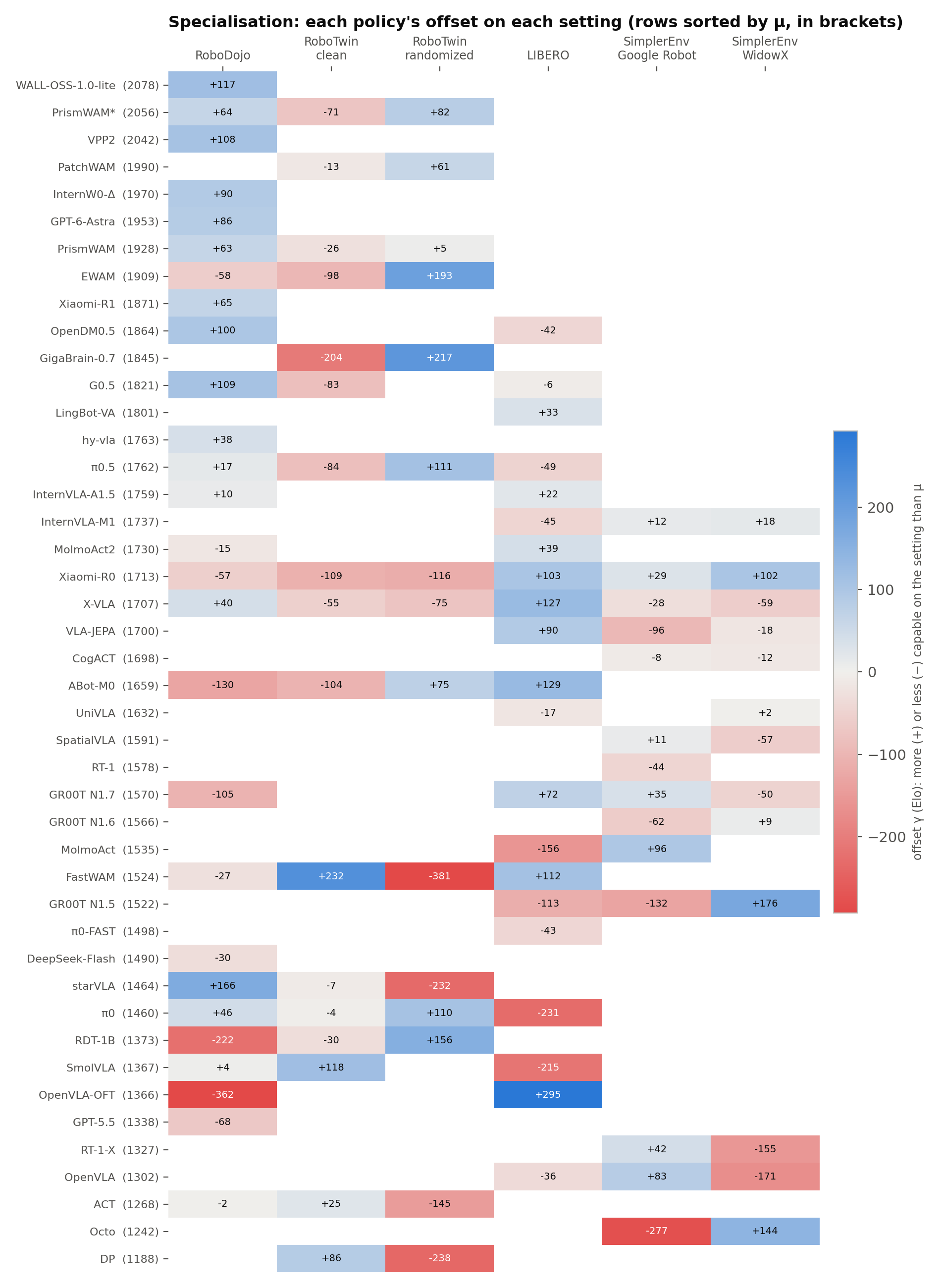}
  \caption{\textbf{Offsets $\gamma$ per policy and setting.} Positive: more capable there than its general capability suggests.}
  \label{fig:offsets}
\end{figure}

\begin{figure}[htbp]
  \centering
  \includegraphics[width=\linewidth]{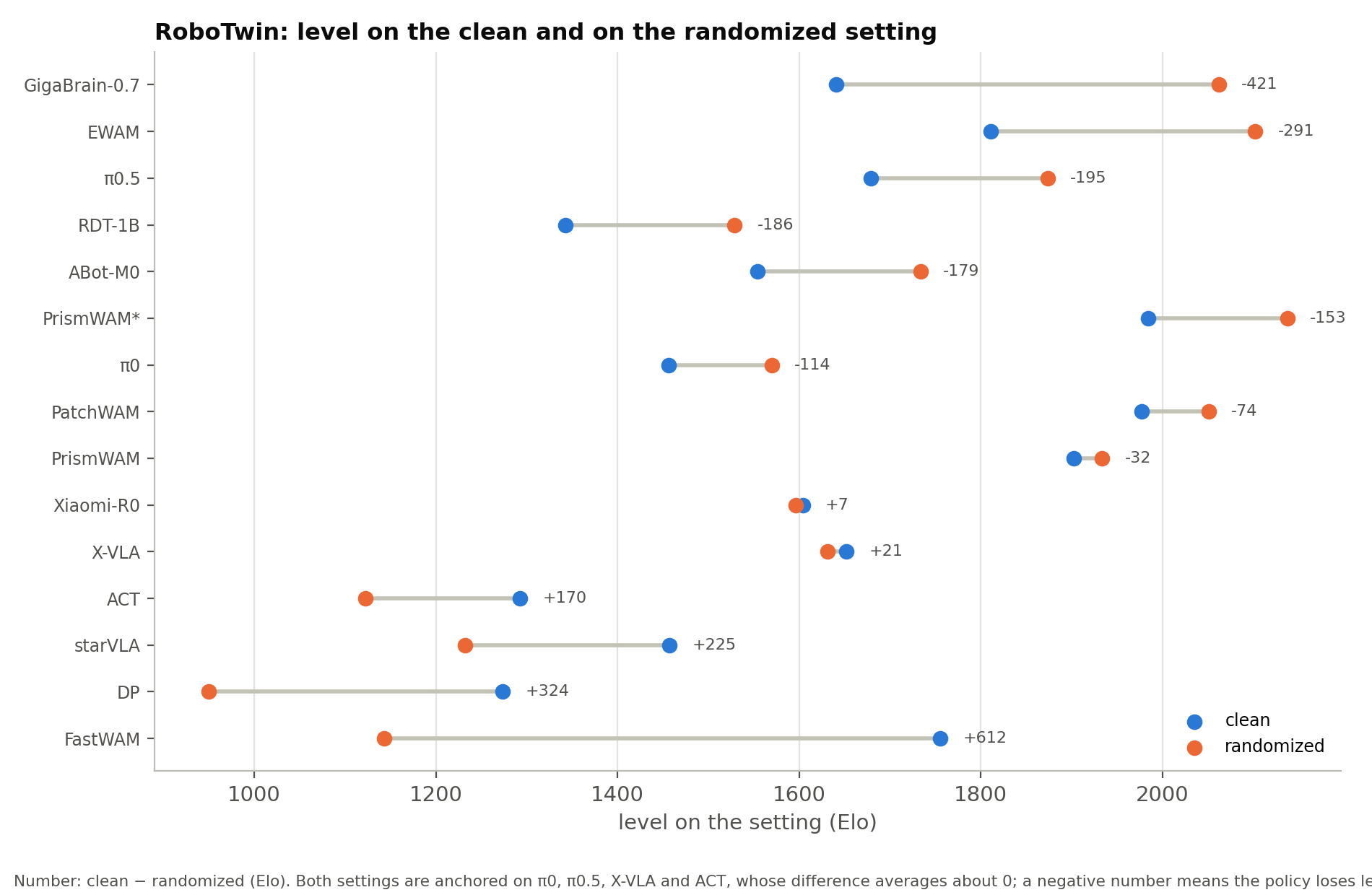}
  \caption{\textbf{RoboTwin~2.0: level on the clean against the randomized setting.} Policies lose very different amounts when the scenes are randomized.}
  \label{fig:robotwin-drop}
\end{figure}

\paragraph{Estimated offsets.}
Figure~\ref{fig:offsets} shows each policy's offsets.
The largest gaps between benchmarks are OpenVLA-OFT (657~Elo points higher on LIBERO than on RoboDojo), SmolVLA (333~Elo points higher on RoboTwin~2.0 than on LIBERO) and RDT-1B (285~Elo points higher on RoboTwin~2.0 than on RoboDojo).
From the clean to the randomized setting of RoboTwin~2.0, policies lose very different amounts: relative to the anchors, the drop ranges from $-421$~Elo points (GigaBrain-0.7) to $+612$~Elo points (FastWAM), with a standard deviation of 262~Elo points (Figure~\ref{fig:robotwin-drop}).
Separate offsets per setting capture such differences: in random-cell cross-validation they lower the held-out log loss on RoboTwin~2.0 from 0.466 to 0.444, and change it little on SimplerEnv (0.542 against 0.541).

\section{Validity and Reliability of the Scale}
\label{sec:reliability}
\label{sec:linking}

\begin{figure}[htbp]
  \centering
  \includegraphics[width=\linewidth]{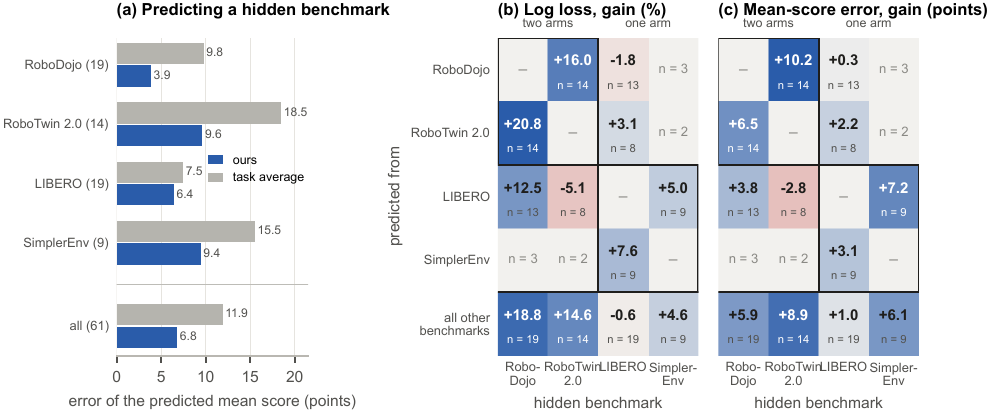}
  \caption{\textbf{The scale predicts hidden benchmarks.} (a)~Error on a hidden benchmark, predicted from all of a policy's other benchmarks; held-out policies in brackets. (b, c)~Which benchmark predicts which: each cell hides the column benchmark and predicts it from the row benchmark alone (bottom row: from all others); gain over the task average. Boxes: two-arm and one-arm benchmarks; grey: fewer than five held-out policies ($n$).}
  \label{fig:prediction}
\end{figure}

\begin{figure}[htbp]
  \centering
  \includegraphics[width=\linewidth]{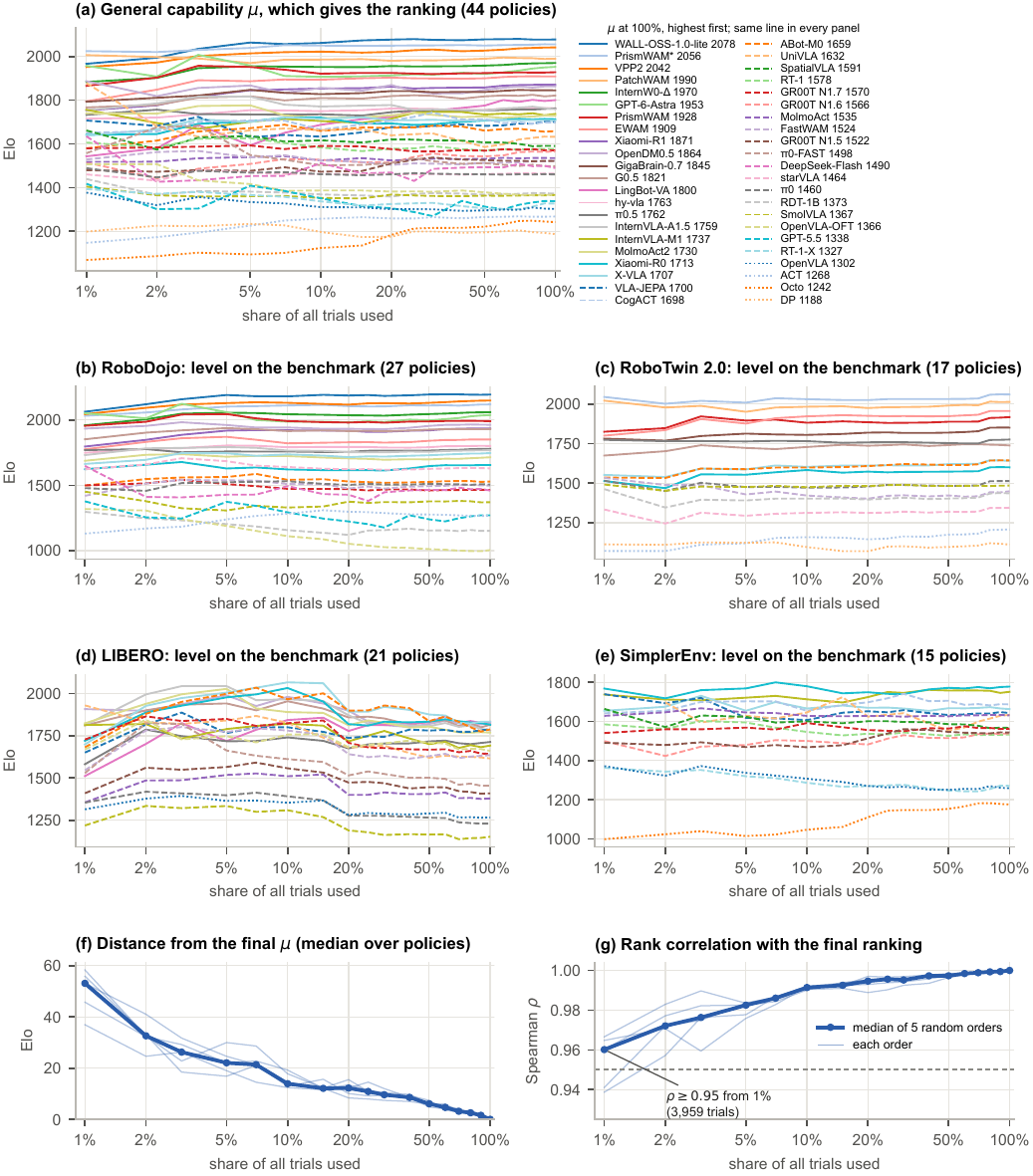}
  \caption{\textbf{Ratings refitted on a growing share of the trials} (one random order; every fit starts from scratch, and 100\% is the main fit). (a)~General capability and (b--e)~level on each benchmark; each policy keeps its line in every panel. (f)~Median distance from the final general capability and (g)~rank correlation with the final ranking, over five random orders.}
  \label{fig:stability}
\end{figure}

\paragraph{Hidden-benchmark prediction.}
For each of the \nLinked{} policies that ran at least two benchmarks, and for each benchmark it ran, we hide all of the policy's results on that benchmark, refit the model, and predict the hidden scores from the policy's other benchmarks.
We compare against the \emph{task average}: the other policies' mean score on the same task, which knows nothing about the hidden policy.
Our predictions are closer on all four benchmarks (Figure~\ref{fig:prediction}a): the error in mean score is 3.9 points against 9.8 on RoboDojo, 9.6 against 18.5 on RoboTwin~2.0, 6.4 against 7.5 on LIBERO, and 9.4 against 15.5 on SimplerEnv, and 6.8 against 11.9 over all 61 hidden cases.
It also orders two hidden policies correctly 71\% to 85\% of the time, while the task average cannot order them at all (Appendix~\ref{app:linking}).

\paragraph{Prediction across benchmark groups.}
Figure~\ref{fig:prediction}b and c repeat the test with a single source benchmark.
The two two-arm benchmarks predict each other well: from RoboDojo, the error on RoboTwin~2.0 falls by 10.2 points.
So do the two one-arm benchmarks.
Across the groups the links are weak: LIBERO even makes predictions for RoboTwin~2.0 worse than the task average, and RoboDojo's predictions for LIBERO are worse in log loss, though not in mean score (Table~\ref{tab:sources}).
The two groups differ in robots and tasks (one arm against two), and benchmark-specific fine-tuning can make a policy far more capable on one benchmark than on another: OpenVLA-OFT, for example, sits 657~Elo points higher on LIBERO than on RoboDojo (Appendix~\ref{app:linking}).
Comparing a policy seen only in one group with one seen only in the other is therefore less certain.

\paragraph{Stability under subsampling.}
We refit the model on a random share of all trials and compare the ranking with the full one (Figure~\ref{fig:stability}g); the ratings themselves settle as in Figure~\ref{fig:stability}a--e.
With \stableShare{} of the trials (\stableTrials{}), the rank correlation already reaches 0.95.
Within each benchmark, RoboDojo and RoboTwin~2.0 settle with 1\% of the trials, while SimplerEnv needs 20\% and LIBERO 50\% (Appendix Figure~\ref{fig:budget}): these two benchmarks hold fewer trials per policy (about 2,700 and 1,900, against 5,700 on RoboDojo and 9,400 on RoboTwin~2.0), and on LIBERO most trials are successes that barely separate the policies.
A new policy can therefore be placed on the scale from a fraction of the usual evaluation.

\paragraph{Additional checks.}
Within each benchmark, ratings agree with raw mean scores (Spearman 0.91 to 1.00); the fit recovers the true capabilities from simulated data (Spearman 0.997, error 14~Elo points); and other model settings or reference panels barely move the ranking (Appendix~\ref{app:checks}).
A model in the style of the Rosetta Stone, fitted to benchmark averages, predicts hidden mean scores about as well (6.4 against our 6.8 points over all hidden cases; Appendix~\ref{app:linking}); what our framework adds is a difficulty for every task and a prediction for every task.

\section{Hidden-Benchmark Evaluation: Full Results}
\label{app:linking}

\paragraph{Full results.}
Table~\ref{tab:linking} reports the test of Appendix~\ref{sec:reliability} in full.
Log loss is per hidden trial; lower is better.
Our pooled log loss is lower than the task average's on RoboDojo, RoboTwin~2.0 and SimplerEnv.
On LIBERO it is slightly higher (0.249 against 0.248), although 74\% of the hidden policies are predicted better and the error in mean score is lower (6.4 against 7.5 points): a few large misses weigh heavily there (see below).
Figure~\ref{fig:linking-policy} shows the gain for each hidden policy.

\begin{table}[htbp]
  \centering
  \caption{\textbf{Hidden-benchmark test.} Error: mean absolute error in a hidden policy's mean score, in points. Order: share of hidden-policy pairs ordered correctly. Rosetta-style: a model fitted to benchmark averages \citep{ho2025rosetta}.}
  \label{tab:linking}
  \small
  \begin{tabular}{@{}lrrrrrrr@{}}
    \toprule
    & & \multicolumn{2}{c}{Log loss} & \multicolumn{3}{c}{Error (points)} & \\
    \cmidrule(lr){3-4}\cmidrule(lr){5-7}
    Hidden benchmark & Policies & Ours & Task avg. & Ours & Rosetta-style & Task avg. & Order \\
    \midrule
    \csname @@input\endcsname sections/tab_linking.tex
    \bottomrule
  \end{tabular}
\end{table}

\paragraph{Single-source prediction.}
Table~\ref{tab:sources} keeps only one source benchmark for the hidden policy.
Pairs with fewer than five shared policies are left out.

\begin{table}[htbp]
  \centering
  \caption{\textbf{Predicting a hidden benchmark from one source benchmark.} Error in points.}
  \label{tab:sources}
  \small
  \begin{tabular}{@{}lrrrr@{}}
    \toprule
    Source $\rightarrow$ hidden & Policies & Error, ours & Error, task avg. & Log loss better? \\
    \midrule
    RoboDojo $\rightarrow$ RoboTwin 2.0 & 14 & 8.3 & 18.5 & yes \\
    RoboTwin 2.0 $\rightarrow$ RoboDojo & 14 & 3.6 & 10.1 & yes \\
    LIBERO $\rightarrow$ SimplerEnv & 9 & 8.3 & 15.5 & yes \\
    SimplerEnv $\rightarrow$ LIBERO & 9 & 3.0 & 6.1 & yes \\
    RoboTwin 2.0 $\rightarrow$ LIBERO & 8 & 7.5 & 9.7 & yes \\
    RoboDojo $\rightarrow$ LIBERO & 13 & 7.6 & 7.9 & no \\
    LIBERO $\rightarrow$ RoboDojo & 13 & 4.8 & 8.6 & yes \\
    LIBERO $\rightarrow$ RoboTwin 2.0 & 8 & 12.9 & 10.1 & no \\
    \bottomrule
  \end{tabular}
\end{table}

\paragraph{Largest prediction errors.}
The largest misses all involve policies that are much more or less capable on the hidden benchmark than on their others (Table~\ref{tab:misses}).
For OpenVLA-OFT the cause is benchmark-specific fine-tuning: it almost never succeeds on RoboDojo, but its LIBERO checkpoint, fine-tuned for LIBERO, nearly solves LIBERO.

\begin{table}[htbp]
  \centering
  \caption{\textbf{The largest misses of the hidden-benchmark test.} Mean scores in percent.}
  \label{tab:misses}
  \small
  \begin{tabular}{@{}llrrr@{}}
    \toprule
    Policy & Hidden benchmark & Observed & Ours & Task average \\
    \midrule
    OpenVLA-OFT & LIBERO & 97.2 & 58.3 & 93.0 \\
    MolmoAct & SimplerEnv & 75.2 & 46.3 & 62.7 \\
    SmolVLA & LIBERO & 67.4 & 92.2 & 94.5 \\
    SmolVLA & RoboTwin 2.0 & 49.6 & 26.9 & 63.7 \\
    \bottomrule
  \end{tabular}
\end{table}

\paragraph{Comparison with an aggregate-score model.}
We also fitted a Rosetta-style model \citep{ho2025rosetta}, $\sigma(\alpha_u (C_m - D_u))$ per setting $u$, to each policy's mean score on each setting.
Over all 61 hidden cases its error is 6.4 points, against 6.8 for ours and 11.9 for the task average.
It is better on RoboTwin~2.0 and LIBERO, and ours is better on RoboDojo and SimplerEnv (Table~\ref{tab:linking}).
The two approaches are close in how well they predict averages; our per-task model additionally gives the difficulty of each task and the expected score on each task.

\begin{figure}[htbp]
  \centering
  \includegraphics[width=\linewidth]{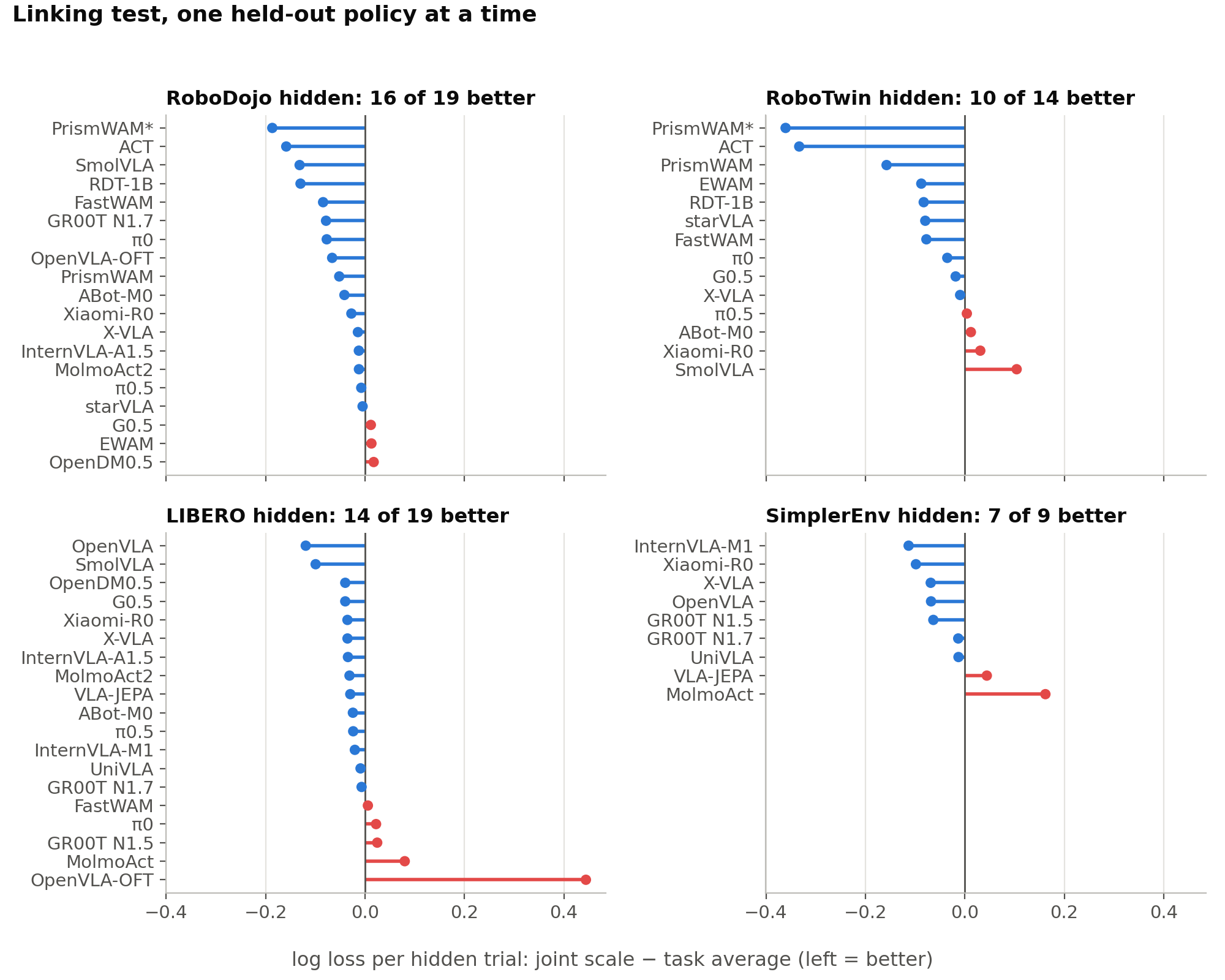}
  \caption{\textbf{Hidden-benchmark test for each hidden policy.} Change in log loss against the task average; negative (blue) is better, and panel titles count such policies.}
  \label{fig:linking-policy}
\end{figure}

\section{Robustness Analyses}
\label{app:checks}

\begin{figure}[htbp]
  \centering
  \includegraphics[width=\linewidth]{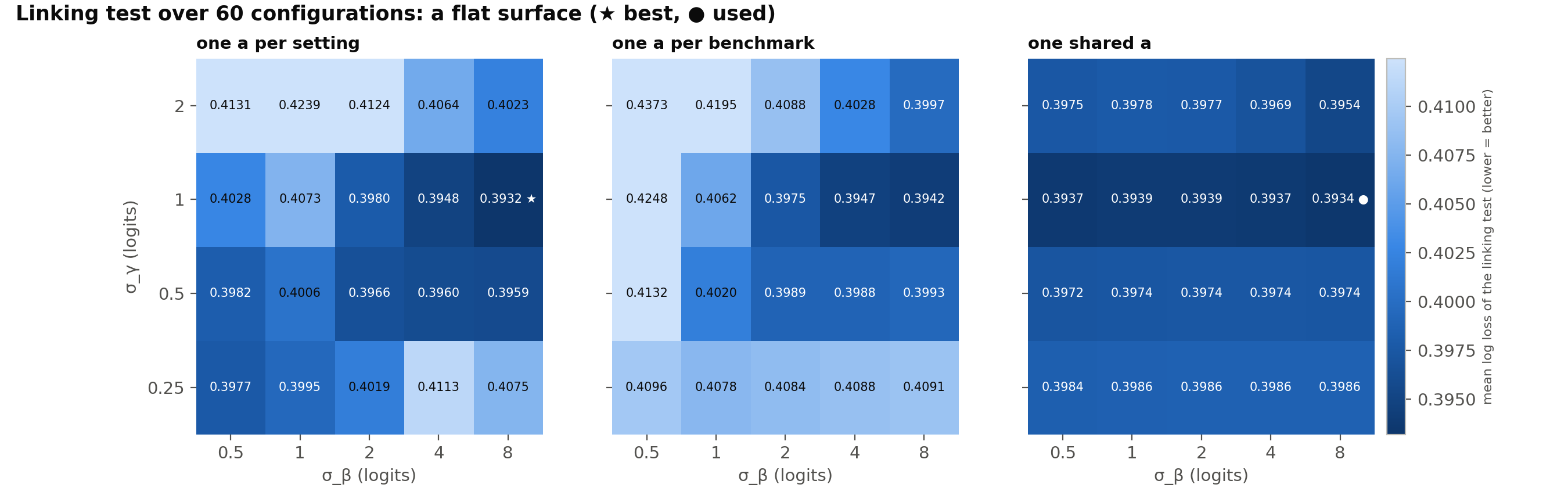}
  \caption{\textbf{Hidden-benchmark test over 60 model settings} (mean log loss; lower is better).}
  \label{fig:grid}
\end{figure}

\begin{figure}[htbp]
  \centering
  \includegraphics[width=\linewidth]{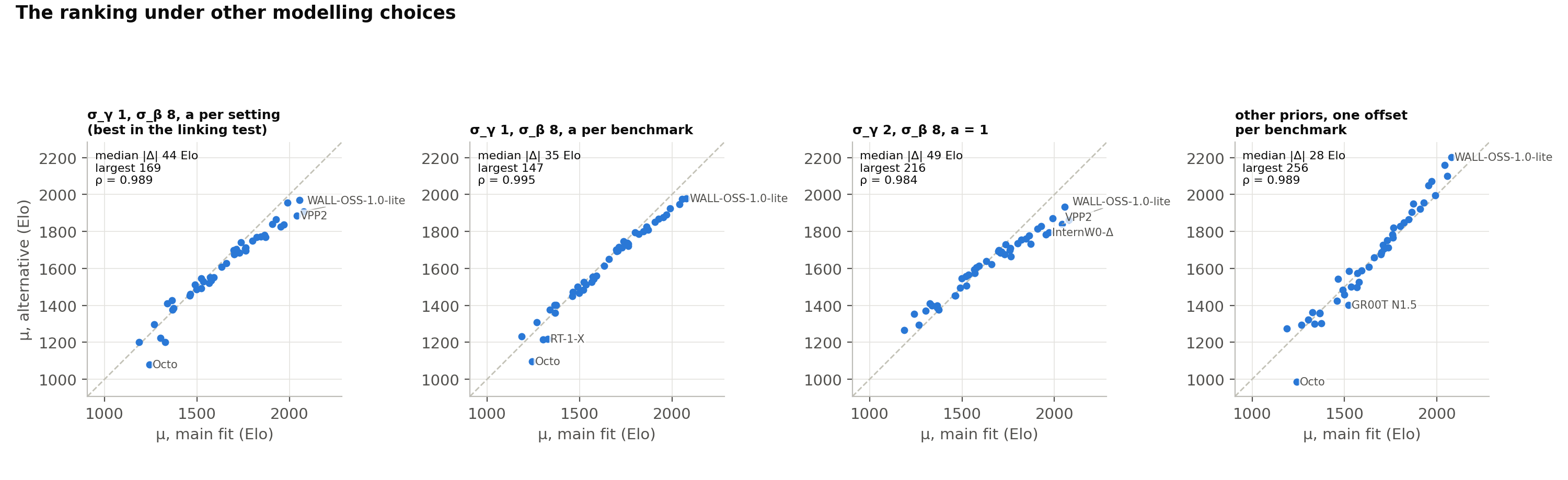}
  \caption{\textbf{General capability under other model settings} against the main fit.}
  \label{fig:robustness}
\end{figure}

\paragraph{Model settings.}
We ran the hidden-benchmark test for 60 settings: $\sigma_\gamma \in \{0.25, 0.5, 1, 2\}$ logits, $\sigma_\beta \in \{0.5, 1, 2, 4, 8\}$ logits, and one shared unit or one conversion factor $a$ per benchmark or per setting, $\log P_{mt}/(1 - P_{mt}) = a\,(\mu_m + \gamma_{ms} - \beta_t)/c$ (Figure~\ref{fig:grid}).
With conversion factors, the likelihood depends only on their product with the ratings, so we fix their geometric mean to one inside the fit; the prior widths then mean the same in every setting.
Ours, with one shared unit, has a mean log loss of \linkLossShared{} and ranks second; the best, with one factor per setting and the same priors, is only 0.0002 lower (\linkLossPerSetting{}), and one factor per benchmark ranks seventh (\linkLossPerBench{}).
One factor per test case predicts worse still: 0.3954 here, and 0.3646 against \cvLossShared{} in random-cell cross-validation.
We keep one shared unit, because the gain from factors is negligible and they would give a gap in Elo a different meaning on each setting.
Under the alternatives of Figure~\ref{fig:robustness}, general capabilities change by a median of 28 to 49~Elo points, with a rank correlation of at least 0.98.
The largest changes concern policies seen on only one benchmark, whose split between $\mu$ and $\gamma$ depends on $\sigma_\gamma$.

\paragraph{Reference panel.}
Changing the reference panel leaves the ranking by $\mu$ exactly unchanged: the anchoring shifts only offsets and difficulties, and the zero point shifts every $\mu$ by the same amount.
Overall scores change by a median of at most 17~Elo points across eight alternative panels.

\paragraph{Policies on one benchmark.}
Refitting on only the \nLinked{} policies with at least two benchmarks changes the remaining capabilities by a median of 6~Elo points (rank correlation 0.998).

\begin{figure}[htbp]
  \centering
  \includegraphics[width=\linewidth]{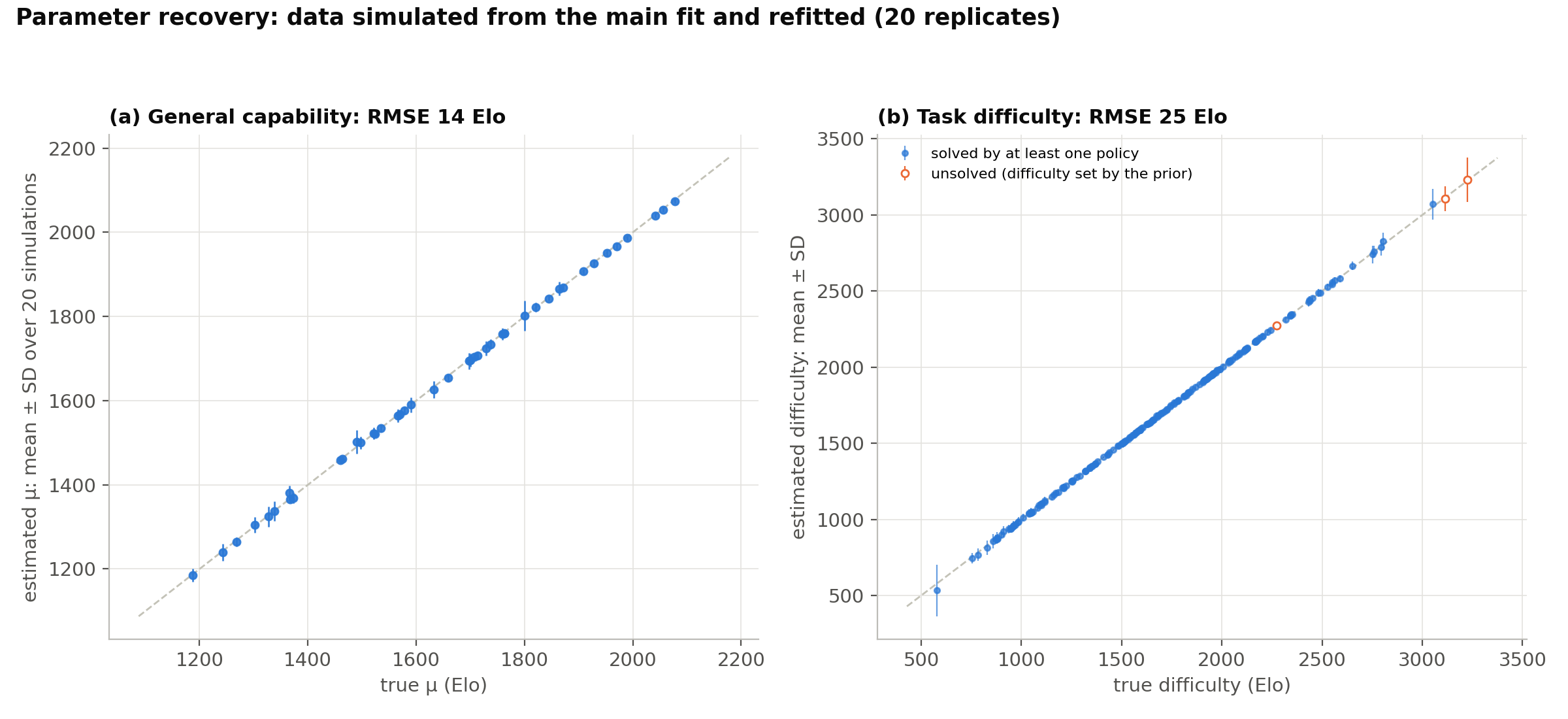}
  \caption{\textbf{Parameter recovery on simulated data:} true against estimated values.}
  \label{fig:recovery}
\end{figure}

\paragraph{Parameter recovery.}
We simulated 20 data sets from the fitted model, with the same cells and trial counts, and refitted each (Figure~\ref{fig:recovery}).
The fit recovers the true capabilities with a rank correlation of $0.997 \pm 0.001$ and an error of $14 \pm 2$~Elo points, and the task difficulties with an error of 25~Elo points.

\begin{figure}[htbp]
  \centering
  \includegraphics[width=\linewidth]{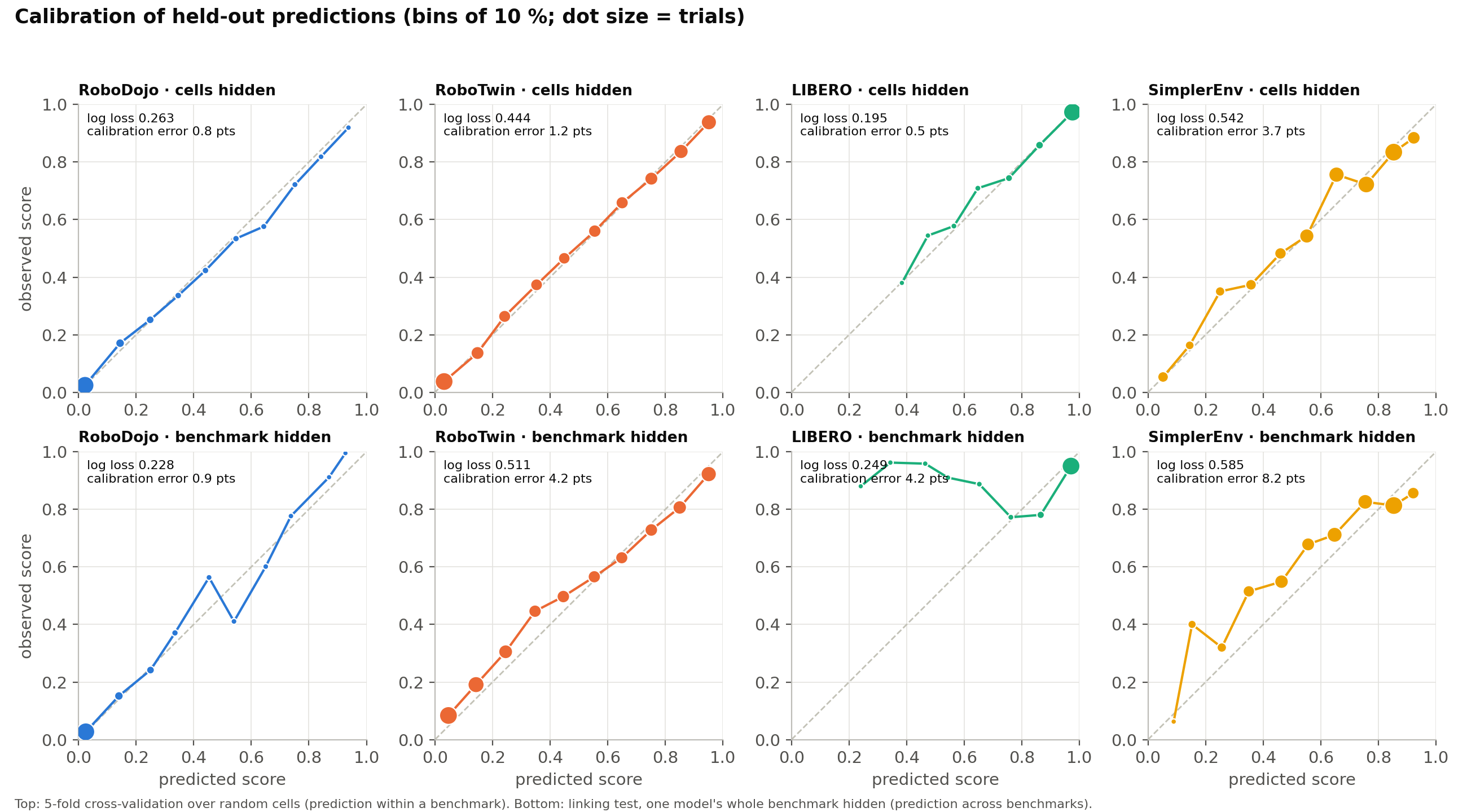}
  \caption{\textbf{Calibration} on random held-out cells and on hidden benchmarks.}
  \label{fig:calibration}
\end{figure}

\paragraph{Calibration and over-dispersion.}
Predicted and observed scores agree on held-out cells, with calibration errors from 0.5 to 3.7 points for random cells and from 0.9 to 8.2 points when a whole benchmark is hidden (Figure~\ref{fig:calibration}).
The residuals are over-dispersed, especially on SimplerEnv and RoboTwin~2.0, so CIs based on the likelihood alone would be too narrow.
Our CIs therefore come from resampling test cases, and they should still be read as lower bounds on the true uncertainty.
The 95\% CI of $\mu$ has a median width of 88~Elo points; it narrows from 136~Elo points for policies on one benchmark to about 70~Elo points for policies on three or four.

\paragraph{Binary success outcomes.}
Fitting binary success instead of partial-success scores changes general capabilities by a median of 18~Elo points (largest 78~Elo points, VPP2; rank correlation 0.996).
WALL-OSS-1.0-lite still leads, with 2155~Elo points; all policies rise by 183~Elo points per year, and the hidden-benchmark error is 7.4 points against 11.5 for the task average.
Partial credit thus changes the details but none of the conclusions.

\paragraph{Third-party re-evaluations.}
Adding the SimplerEnv results of the Manipulation Benchmark Audit~\citep{jiang2026benchmarking}, which use shifted test conditions, changes capabilities by a median of 3~Elo points after aligning the scale.

\begin{figure}[htbp]
  \centering
  \includegraphics[width=\linewidth]{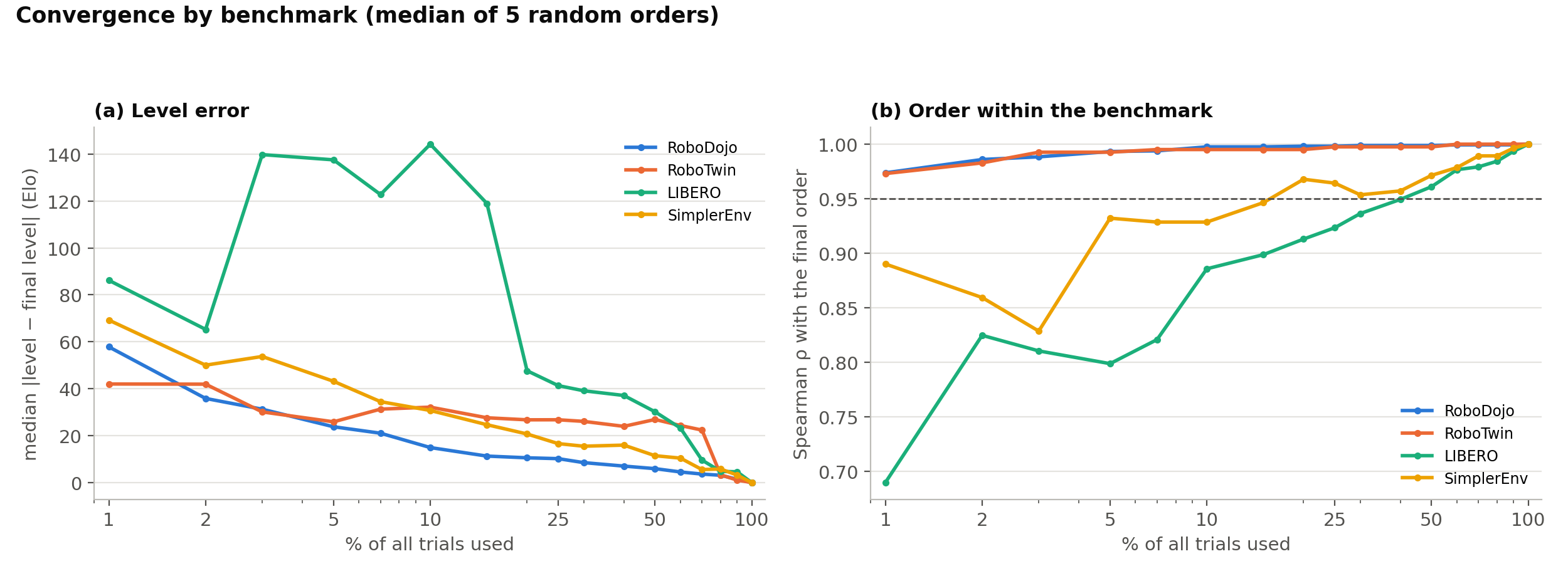}
  \caption{\textbf{Trials needed per benchmark:} change in level (a) and agreement with the final order within the benchmark (b) against the share of all trials used; median of five random orders.}
  \label{fig:budget}
\end{figure}

\paragraph{Evaluation budget per benchmark.}
Figure~\ref{fig:budget} repeats the test of Figure~\ref{fig:stability}g for the order of the policies within each benchmark.
RoboTwin~2.0 and RoboDojo reach a rank correlation of 0.95 with 1\% of all trials; SimplerEnv and LIBERO need 20\% and 50\%. They hold fewer trials per policy (about 2,700 and 1,900, against 5,700 on RoboDojo and 9,400 on RoboTwin~2.0), so a given share of all trials leaves them fewer, and on LIBERO most trials are successes that barely separate the policies.

\section{Trend Estimation and Forecasting}
\label{app:trend}

\paragraph{Change-point selection.}
The SOTA line of Figure~\ref{fig:teaser} bends once, at G0.5 (16 June 2026), because the record rose at very different paces before and after it: 69~Elo points per year from RT-1 to G0.5 and 727 from G0.5 to WALL-OSS-1.0-lite.
To choose the kink, we placed it on each of the ten record setters between RT-1 and WALL-OSS-1.0-lite in turn and fitted all twelve record setters in two ways: the line through RT-1, the kink and WALL-OSS-1.0-lite, as drawn in Figure~\ref{fig:teaser}, and a continuous two-segment least-squares line with its kink on that date (Table~\ref{tab:kink}).
G0.5 fits best under both (root-mean-square error 14.3 and 11.5~Elo points, against 71.4 for one straight line) and is the best kink in 48\% and 59\% of the 1,000 bootstrap fits.
OpenDM0.5 and Xiaomi-R1, released within a month after G0.5, fit almost as well, take most of the remaining fits and give similar paces after the kink (738 and 764~Elo points per year); a kink at LingBot-VA, the first WAM in our data, or at GPT-6-Astra, the first LLM Agent to set a record, fits much worse.
The change point therefore lies in June or July 2026, and G0.5 is its best single estimate.
Each period's pace runs from its first record to the first record of the next (for the second period, to its last record).

\begin{table}[htbp]
  \centering
  \caption{\textbf{Where the SOTA line bends.} Error of a SOTA line with one kink, for each record setter between RT-1 and WALL-OSS-1.0-lite as the kink: root-mean-square error over the twelve record setters (Elo points) and the share of the 1,000 bootstrap fits (tasks resampled) in which it is the best kink. Anchored: the line through RT-1, the kink and WALL-OSS-1.0-lite, as in Figure~\ref{fig:teaser}; least squares: a continuous two-segment line. One straight line: 71.4.}
  \label{tab:kink}
  \small
  \begin{tabular}{@{}llrrrr@{}}
    \toprule
    & & \multicolumn{2}{c}{Anchored} & \multicolumn{2}{c}{Least squares} \\
    \cmidrule(lr){3-4} \cmidrule(l){5-6}
    Kink at & Released & RMSE & Best (\%) & RMSE & Best (\%) \\
    \midrule
    CogACT               & 2024-11 & 98.1  & 0.0  & 61.2 & 0.0  \\
    $\pi_{0.5}$          & 2025-04 & 96.4  & 0.0  & 59.4 & 0.0  \\
    LingBot-VA           & 2026-01 & 62.4  & 0.2  & 42.2 & 0.0  \\
    \textbf{G0.5}        & 2026-06 & \textbf{14.3} & \textbf{48.0} & \textbf{11.5} & \textbf{58.6} \\
    OpenDM0.5            & 2026-07 & 20.6  & 27.9 & 13.2 & 19.5 \\
    Xiaomi-R1            & 2026-07 & 22.1  & 17.0 & 14.2 & 12.3 \\
    GPT-6-Astra          & 2026-09 & 57.5  & 6.9  & 29.7 & 9.6  \\
    PatchWAM             & 2026-09 & 76.1  & 0.0  & 41.5 & 0.0  \\
    VPP2                 & 2026-10 & 105.4 & 0.0  & 53.0 & 0.0  \\
    PrismWAM$^{*}$       & 2026-10 & 111.8 & 0.0  & 62.5 & 0.0  \\
    \bottomrule
  \end{tabular}
\end{table}

\paragraph{Estimated trends.}
Table~\ref{tab:trends} collects the trends quoted in Section~\ref{sec:results}, each with the time in which it doubles the odds of success on a task.

\begin{table}[htbp]
  \centering
  \caption{\textbf{Trends on the Elo scale.} In brackets: 95\% CI (tasks and policies resampled; RoboDojo: tasks resampled; benchmark medians: least-squares CI of the slope over the five benchmarks, as in the abstract). $R^2$: of each line over the points it describes. Doubling time: months per 120~Elo points, the gain that doubles the odds of success on a task.}
  \label{tab:trends}
  \small
  \begin{tabular}{@{}lrrrr@{}}
    \toprule
    Trend & $n$ & Elo points per year & $R^2$ & Doubling time \\
    \midrule
    All policies                 & 44 & 171 (116--238)  & 0.55 & 8.5 months \\
    \quad Before G0.5 (to June 2026)       & 30 & 107 (39--178)   & 0.36 & 13.5 months \\
    \quad Since G0.5                       & 14 & 608 (194--1004) & 0.24 & 2.4 months \\
    Record setters               & 12 & 117 (89--264)   & 0.77 & 12.3 months \\
    \quad Before G0.5 (to June 2026)       & 4 & 69 (54--232)     & 0.97 & 20.8 months \\
    \quad Since G0.5                       & 8 & 727 (536--1323)  & 0.97 & 2.0 months \\
    Records, last six months     & 8  & 555             & --   & 2.6 months \\
    Benchmark medians            & 5  & 372 (120--624)  & 0.88 & 3.9 months \\
    Best level on RoboDojo       & 6  & 877 (614--1259) & 0.88 & 1.6 months \\
    \bottomrule
  \end{tabular}
\end{table}

\paragraph{Linear versus exponential trends.}
We compare a straight line with a curve that rises exponentially in Elo, $c + A e^{r(t - 2026)}$, by their leave-one-out error (Table~\ref{tab:trend}).
The curve fits better for all policies, for the policies that ran a two-arm benchmark, and without RT-1, whose early rating rests on SimplerEnv alone; for the policies released from 2024 on, the two are tied.
The curve does not fit better once the policies released after June 2026 are removed: the speed-up rests mostly on the latest policies.
For all policies, the curve's growth rate $r$ is positive in 98\% of bootstrap samples (tasks and policies resampled).

\begin{table}[htbp]
  \centering
  \caption{\textbf{Line against curve.} Slope of the line in Elo points per year; leave-one-out (LOO) error in Elo points.}
  \label{tab:trend}
  \small
  \begin{tabular}{@{}lrrrr@{}}
    \toprule
    Policies & $n$ & Line slope & LOO, line & LOO, curve \\
    \midrule
    All & 44 & 171 & 170 & \textbf{156} \\
    Without RT-1 & 43 & 200 & 153 & \textbf{149} \\
    Ran RoboDojo or RoboTwin 2.0 & 30 & 206 & 173 & \textbf{159} \\
    LIBERO or SimplerEnv only & 14 & 104 & 181 & \textbf{177} \\
    Released from 2024 on & 40 & 231 & 155 & 155 \\
    Released before July 2026 & 31 & 113 & \textbf{154} & 158 \\
    \bottomrule
  \end{tabular}
\end{table}

\paragraph{Retrospective forecasts.}
A straight line fitted to the policies released before mid-2025 falls short of the 28 later policies by 140~Elo points on average; before January 2026, of the 24 later policies by 106~Elo points; and before April 2026, of the 17 later policies by 162~Elo points.
A line through the record setters released before each cut-off also falls short of the later record setters, by 102 to 116~Elo points on average, so the speed-up shows in the sequence of records as well as in the bulk of policies.

\begin{figure}[htbp]
  \centering
  \includegraphics[width=\linewidth]{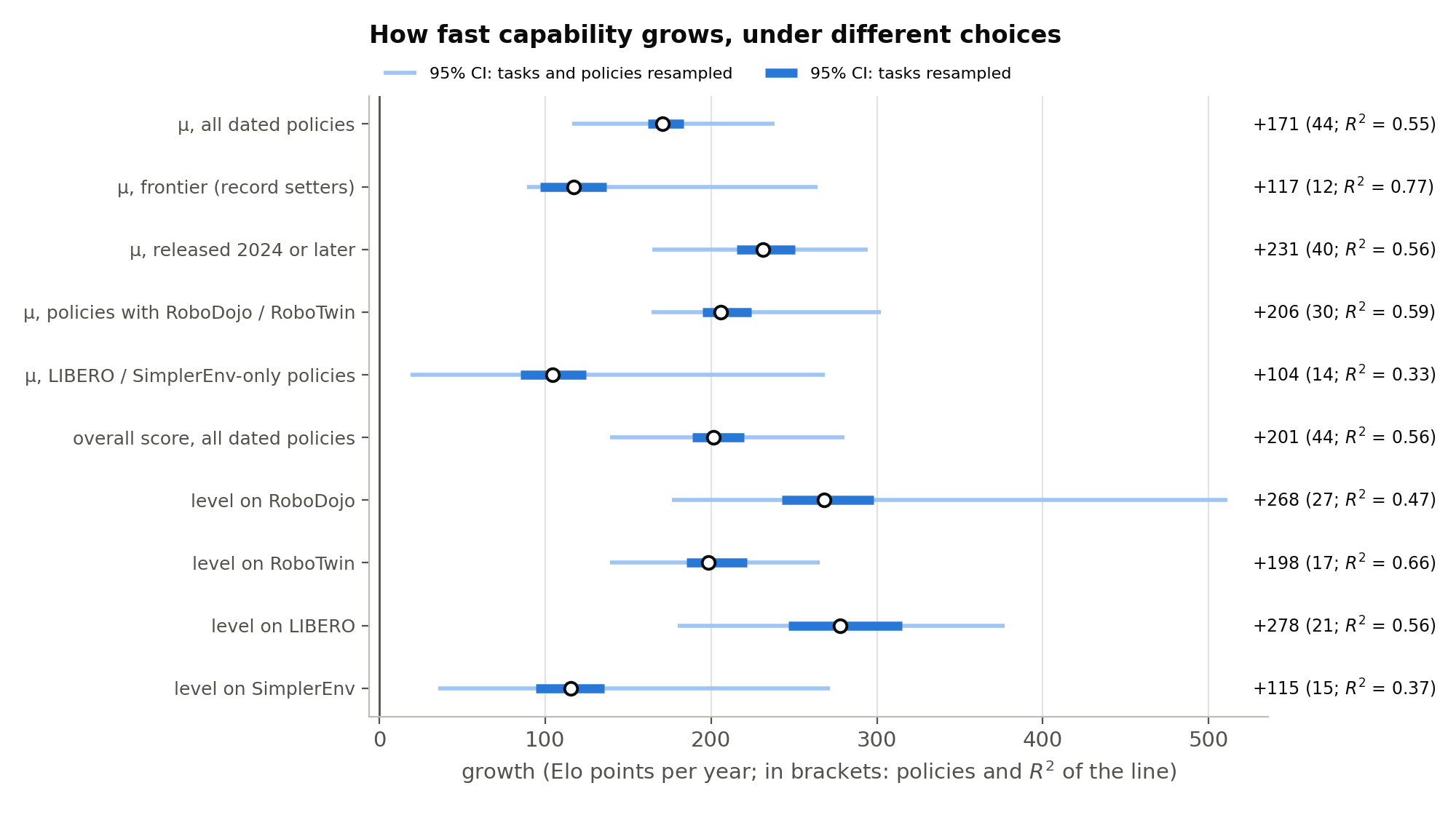}
  \caption{\textbf{Growth rate under other choices} of policies and scores, with 95\% CIs and the $R^2$ of each line.}
  \label{fig:growth}
\end{figure}

\paragraph{Sensitivity of the growth rate.}
Using the overall score instead of $\mu$ gives a slope of 201~Elo points per year.
Each benchmark on its own rises by 268 (RoboDojo), 198 (RoboTwin~2.0), 278 (LIBERO) and 115 (SimplerEnv)~Elo points per year (Figure~\ref{fig:growth}).
Because the way a scale is built can change the shape of growth curves \citep{briggs2009impact}, we repeated the analysis with the overall scores of all 60 model settings of Appendix~\ref{app:checks}.
The slope ranges from 197 to 203~Elo points per year with one shared unit and from 54 to 203 over all 60 settings, the lowest coming from one conversion factor per setting with a narrow prior on task difficulty; the curve fits better in 59 of the 60 settings, and the policies released after each back-test cut-off lie above the line in all 60.

\paragraph{Saturation forecast from RoboDojo levels.}
A question about one benchmark uses levels on it, $\mu_m + \gamma_{ms}$.
We call a benchmark \emph{saturated} when the policy with the highest level on it has an expected score of at least 50\% on 90\% of its tasks; a result counts from the later of the policy's and the benchmark's release.
Table~\ref{tab:usedup} gives the months each benchmark took.
RoboTwin~2.0 clean was saturated at its release, by $\pi_{0.5}$, so for benchmarks that were hard at their release the range is \usedUpLo{} to \usedUpHi{} months.
LIBERO's 12 months are an upper bound, as the first policy in our data was evaluated on it a year after its release; LIBERO is also the only benchmark on which the best policy has reached an expected score of at least 90\% on 90\% of the tasks, after 21 months.
On RoboDojo the best level rose from \dojoRecordRelease{}~Elo points (G0.5, at its release on 5 July 2026) to 1964 (OpenDM0.5) in July, 2038 (GPT-6-Astra) and 2060 (InternW0-$\Delta$) in September, and 2149 (VPP2) and \dojoRecordNow{} (WALL-OSS-1.0-lite) in October 2026, 877~Elo points per year (95\% CI: 614--1259, tasks resampled; $R^2 = 0.88$).
Extended from the last record, this pace reaches the median task difficulty in November 2026 and the 90th percentile in \dojoNinetyDate{} (95\% CI: \dojoNinetyLo{}--\dojoNinetyHi{}), three months before the main forecast of Section~\ref{sec:left}, which extends general capability at its pace since G0.5 (\forecastNinetyDate{}).
It rests on six policies in less than four months, and the pace on a benchmark can slow after its first months: the best level on RoboTwin~2.0 randomized rose by about 200~Elo points per year in the 16 months after its release, and at that pace RoboDojo's 90th percentile would be reached only in 2029.
The months are counted from the release dates of the policies, not of their leaderboard entries, and our data hold only the \nPolicies{} milestone policies, so Table~\ref{tab:usedup} is approximate.

\begin{table}[htbp]
  \centering
  \caption{\textbf{How fast each benchmark saturated:} months from its release until the policy with the highest level has an expected score of at least 50\% on 90\% of its tasks, and where the best policy stands today.}
  \label{tab:usedup}
  \small
  \begin{tabular}{@{}llrlr@{}}
    \toprule
    Benchmark (setting) & Released & Months to saturation & Best policy today (level) & Tasks $\geq$\,50\% today \\
    \midrule
    LIBERO                     & 2023-06 & 12 (upper bound) & X-VLA (1834)                 & 100\% \\
    SimplerEnv Google Robot    & 2024-05 & 7                & InternVLA-M1 (1749)          & 100\% \\
    SimplerEnv WidowX          & 2024-05 & 21               & Xiaomi-R0 (1815)             & 100\% \\
    RoboTwin 2.0 clean         & 2025-06 & 0 (at release)   & PrismWAM$^{*}$ (1984)        & 100\% \\
    RoboTwin 2.0 randomized    & 2025-06 & 16               & PrismWAM$^{*}$ (2137)        & 96\% \\
    RoboDojo                   & 2026-07 & not yet (4 months old) & WALL-OSS-1.0-lite (2195) & 44\% \\
    \bottomrule
  \end{tabular}
\end{table}

\begin{figure}[htbp]
  \centering
  \includegraphics[width=\linewidth]{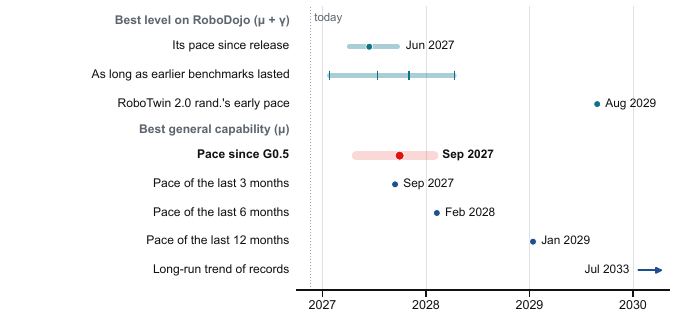}
  \caption{\textbf{When RoboDojo saturates, under other assumptions.} Red: the main forecast (95\% CI); ticks: the lifetimes of earlier benchmarks; arrow: beyond 2030.}
  \label{fig:sensitivity}
\end{figure}

\paragraph{Saturation forecast from general capability.}
The main forecast of Section~\ref{sec:left} is for general capability: it assumes that future policies have zero offset on RoboDojo, so that their level there equals $\mu$.
Since G0.5 (Figure~\ref{fig:teaser}), the record of general capability rose from 1821~Elo points (G0.5, 16 June 2026) to 2078~Elo points (WALL-OSS-1.0-lite, 23 October 2026), 727~Elo points per year (95\% CI: 536--1323, tasks and policies resampled; $R^2 = 0.97$); extended from WALL-OSS-1.0-lite, this pace reaches RoboDojo's median task difficulty in \forecastMedian{} and its 90th percentile in \forecastNinetyDate{} (95\% CI: \forecastNinetyLo{}--\forecastNinetyHi{}).
Paces over longer windows give later dates, for two reasons.
First, they start lower: $\mu$ leaves out gains specific to RoboDojo and pulls the policies seen on RoboDojo alone, among them today's leader, toward the average; the level of WALL-OSS-1.0-lite on RoboDojo is 2195~Elo points, 117~Elo points above its general capability, which gives an expected score of 42\% on the median task against 27\% at zero offset.
Second, they include slower periods; over RoboDojo's first months, the record of general capability itself rose by about 850~Elo points per year, close to the record level on RoboDojo.
The frontier is the sequence of policies that set a new record when released (twelve policies).
A straight line through them rises by 117~Elo points per year (95\% CI: 89--264, tasks and policies resampled; $R^2 = 0.77$) and reaches RoboDojo's median task difficulty (\dojoMedian{}~Elo points) in March 2029.
An exponential curve through the same twelve policies reaches it in January 2027.
The pace of the last six months in Sections~\ref{sec:progress} and~\ref{sec:left} is the gain of the record of general capability over a window that ends in November 2026, extended from today's best policy (WALL-OSS-1.0-lite, 2078~Elo points).
Over the last three, six and twelve months the record rose by 207, 278 and 316~Elo points; at these paces the best policies would reach the median task difficulty (\dojoMedian{}~Elo points) in February, March and June 2027, and the 90th percentile (\dojoNinety{}~Elo points) in September 2027, February 2028 and January 2029.
PrismWAM$^{*}$, whose RoboDojo results come from its own paper, is 22~Elo points behind and held the record for only four days, so these dates, which start from WALL-OSS-1.0-lite, do not depend on it.
These are naive extrapolations: the pace since G0.5 rests on eight records within four months.
We therefore report them only as a range.
Figure~\ref{fig:sensitivity} collects the dates of this section.

\begin{figure}[htbp]
  \centering
  \includegraphics[width=\linewidth]{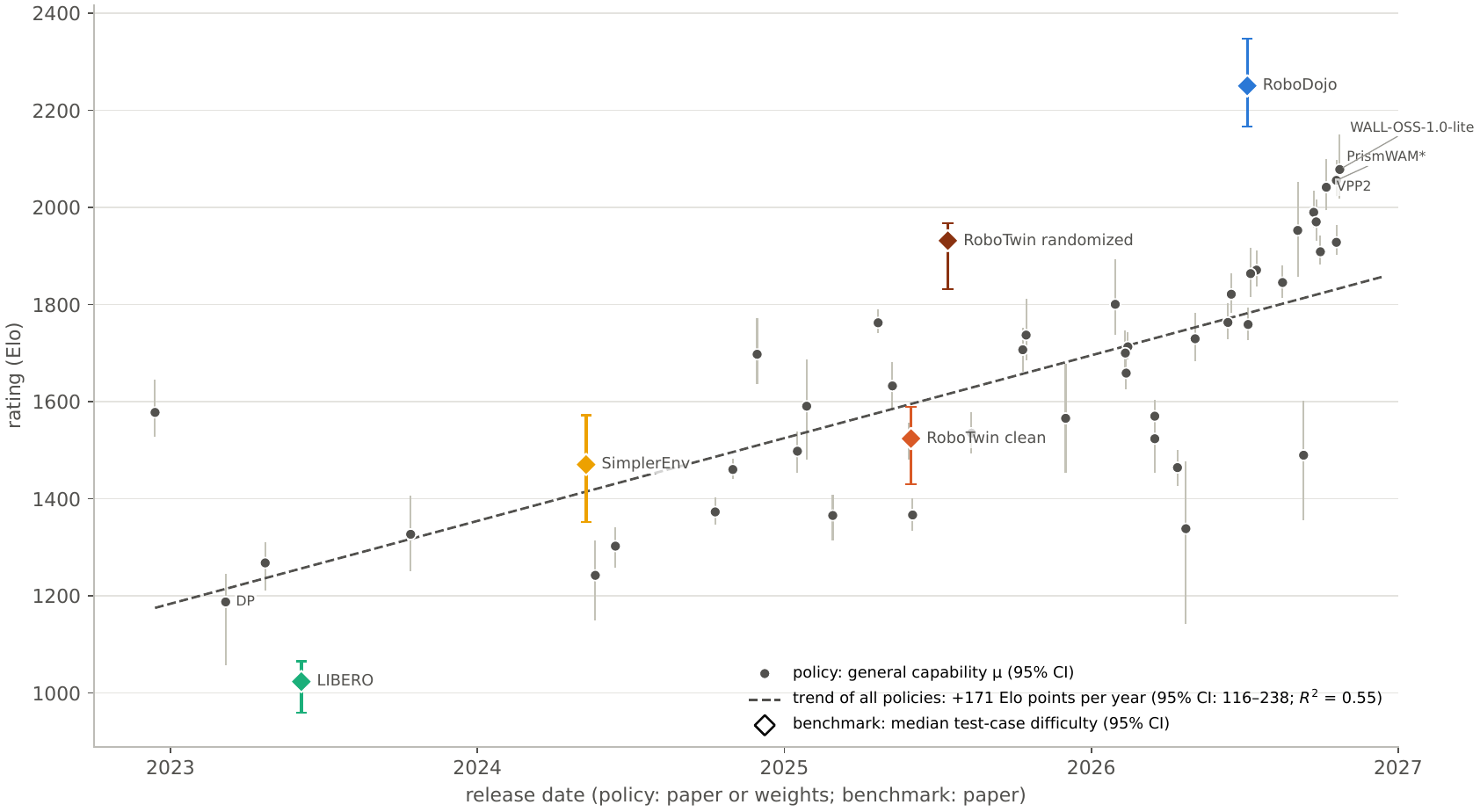}
  \caption{\textbf{General capability of every policy and median task difficulty of every benchmark by release date,} with 95\% CIs (benchmarks: tasks resampled). Dashed: the trend of all policies, as in Figure~\ref{fig:progress}a. RoboTwin~2.0's markers are offset.}
  \label{fig:overtime}
\end{figure}

\paragraph{Capability and difficulty over time.}
Figure~\ref{fig:overtime} shows the data of Figure~\ref{fig:teaser} with 95\% CIs: every policy's general capability and every benchmark's median task difficulty by release date.
A policy whose level on a benchmark equals the benchmark's median difficulty has an expected score of at least 50\% on half of its tasks.
The figure shows general capability, which equals the level only for a policy with zero offset; a policy's level on a benchmark lies above or below its general capability by its offset (Figure~\ref{fig:offsets}).

\begin{figure}[htbp]
  \centering
  \includegraphics[width=\linewidth]{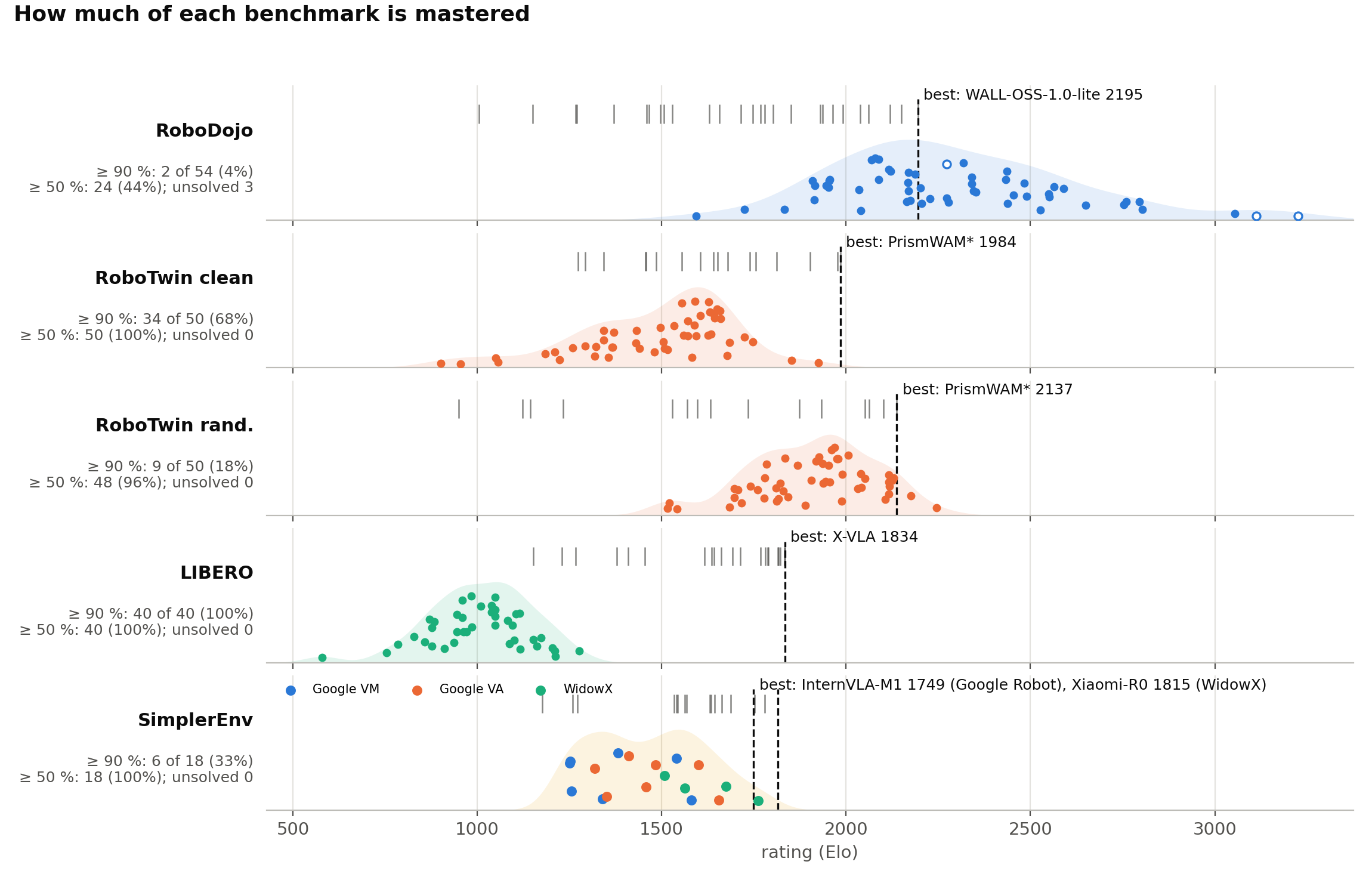}
  \caption{\textbf{Tasks and the best policy on each benchmark.} Dots: task difficulty, spread vertically inside the distribution only for visibility (hollow: no policy ever succeeded); dashed line: the best policy's level; grey ticks: the other policies' levels.}
  \label{fig:mastery}
\end{figure}

\paragraph{Mastery of each benchmark.}
Figure~\ref{fig:mastery} shows, for each benchmark, every task's difficulty and the level of the best policy there, which gives the shares quoted in Section~\ref{sec:left}; Figure~\ref{fig:usedup}a shows the same distributions without the single tasks.

\section{Sensitivity to Evaluation Settings}
\label{app:settings}

\begin{figure}[htbp]
  \centering
  \includegraphics[width=\linewidth]{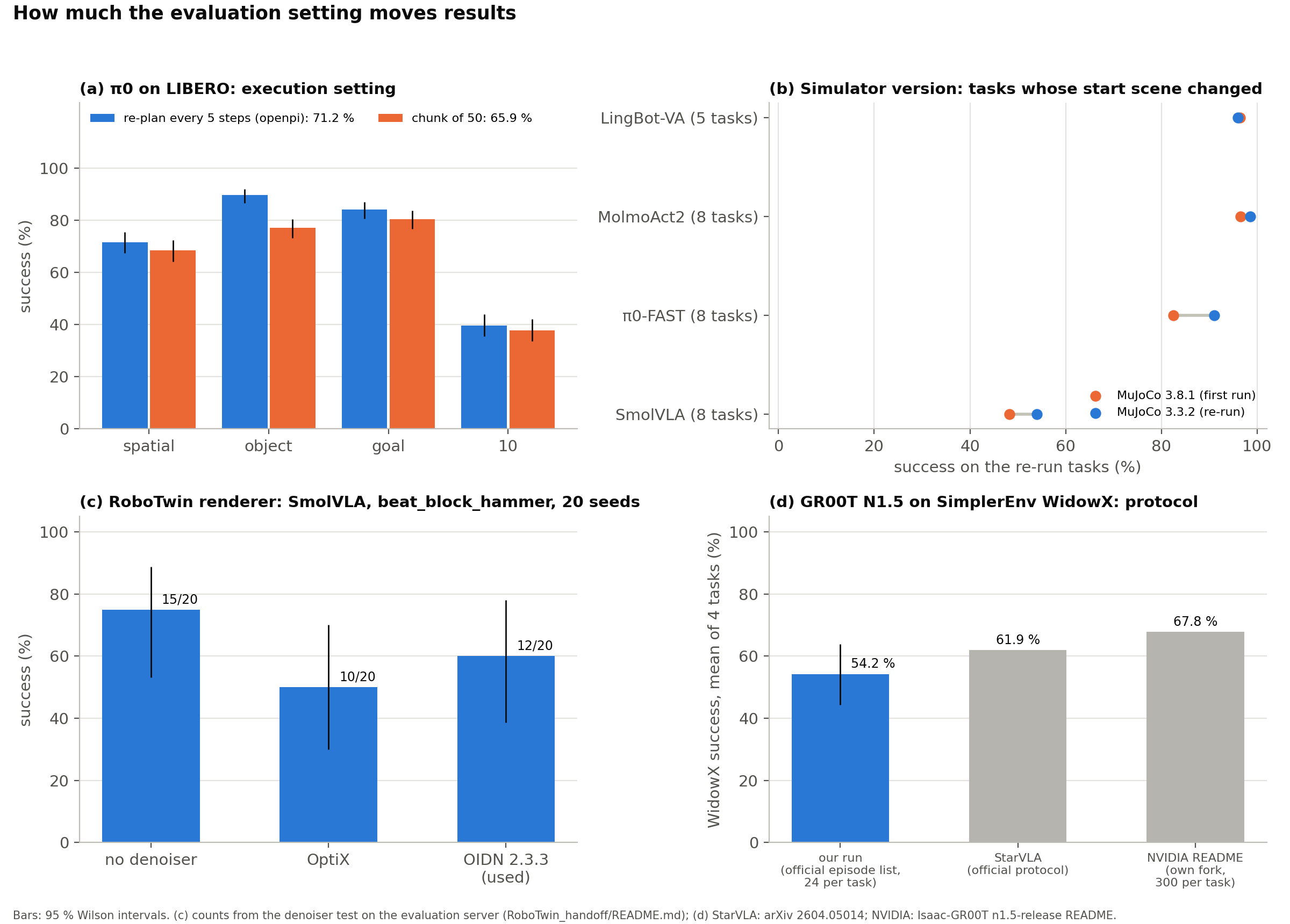}
  \caption{\textbf{The evaluation setting moves results} for the same checkpoint.}
  \label{fig:settings}
\end{figure}

The same checkpoint can score differently under settings that leaderboards rarely report (Figure~\ref{fig:settings}).
\begin{itemize}
  \item \emph{Action chunking.} $\pi_0$ on LIBERO succeeds 71.2\% of the time when it re-plans every five steps (openpi's setting), and 65.9\% when it executes its full chunk of 50 actions.
  \item \emph{Simulator version.} Moving LIBERO from MuJoCo~\citep{todorov2012mujoco} 3.3.2 to 3.8.1 barely changes tasks whose starting scene is unchanged ($\pi_0$: 69.7\% against 70.0\%), but one task whose scene changed went from 11 to 37 successes out of 50 for $\pi_0$, and from 19 to 47 for $\pi_0$-FAST.
  \item \emph{Reported numbers.} For GR00T N1.5 on SimplerEnv's WidowX tasks, our run gives 54.2\% (95\% CI: 44.2--63.8), a re-run by StarVLA~\citep{ye2026starvla} 61.9\%, and NVIDIA's own report~\citep{nvidia2025gr00tn15simpler} 67.8\%.
  \item \emph{Rendering.} On one RoboTwin~2.0 task, SmolVLA succeeded in 15, 10 and 12 of the same 20 seeds without a denoiser, with OptiX~\citep{parker2010optix} and with OIDN~\citep{afra2024oidn}.
\end{itemize}
We fixed and recorded all such settings for our own runs.

\end{document}